\documentclass{acmsmall} 
\pdfoutput=1
\usepackage{graphicx}
\usepackage{caption}
\usepackage{color}
\usepackage{subfigure}
\usepackage{multirow}
\usepackage{url}
\usepackage{graphicx}
\usepackage{times,amsmath}
\usepackage{amsfonts}

\usepackage[ruled]{algorithm2e}

\SetAlFnt{\small}
\SetAlCapFnt{\small}
\SetAlCapNameFnt{\small}
\SetAlCapHSkip{0pt}
\IncMargin{-\parindent}

\setcopyright{rightsretained}

\begin{document}

\markboth{W. Wang et al.}{Deep Learning At Scale and At Ease}

\title{Deep Learning At Scale and At Ease}
\author{Wei Wang
\affil{School of Computing, National University of Singapore, Singapore}
Gang Chen
\affil{College of Computer Science, Zhejiang University, China}
Haibo Chen
\affil{NetEase, Inc., China}
Tien Tuan Anh Dinh and Jinyang Gao and Beng Chin Ooi and \\Kian-Lee Tan and Sheng Wang
\affil{School of Computing, National University of Singapore, Singapore}
}

\begin{abstract}
Recently, deep learning techniques have enjoyed success in various multimedia applications, such as image classification and multi-modal data analysis. Large deep learning models are developed for learning rich representations of complex data. There are two challenges to overcome before deep learning can be widely adopted in multimedia and other applications. One is usability, namely the implementation of different models and training algorithms must be done by non-experts without much effort especially when the model is large and complex. The other is scalability, that is the deep learning system must be able to provision for a huge demand of computing resources for training large models with massive datasets. To address these two challenges, in this paper, we design a distributed deep learning platform called SINGA which has an intuitive programming model based on the common layer abstraction of deep learning models. Good scalability is achieved through flexible distributed training architecture and specific optimization techniques. SINGA runs on GPUs as well as on CPUs, and we show that it outperforms many
other state-of-the-art deep learning systems. Our experience with developing and training deep learning models for real-life multimedia applications in SINGA shows that the platform is both usable and scalable. 
\end{abstract}
%
%


\maketitle

\section{Introduction}
In recent years, we have witnessed successful adoptions of deep learning in various multimedia applications, such as image and video classification~\cite{DBLP:conf/nips/KrizhevskySH12,DBLP:conf/mm/WuJWPX14}, content-based image retrieval~\cite{DBLP:conf/mm/WanWHWZZL14}, music recommendation~\cite{DBLP:conf/mm/WangW14} and multi-modal data analysis~\cite{DBLP:journals/pvldb/WangOYZZ14,DBLP:conf/mm/FengWL14,DBLP:conf/mm/ZhangYLYC14}. Deep learning refers to a set of feature learning models which consist of multiple layers.  Different layers learn different levels of abstractions (or features) of the raw input data~\cite{DBLP:conf/icml/LeRMDCCDN12}.  It has been regarded as a re-branding of neural networks developed twenty years ago, since it inherits many key neural networks techniques and algorithms.  However, deep learning exploits the fact that high-level abstractions are better at representing the data than raw, hand-crafted features, thus achieving better performance in learning.  Its recent resurgence is mainly fuelled by higher than ever accuracy obtained in image recognition~\cite{DBLP:conf/nips/KrizhevskySH12}.  Three key factors behind deep learning's remarkable achievement are the advances of neural net structures, immense computing power and the availability of massive training datasets, which together enable us to train large models to capture the regularities of complex data more efficiently than twenty years ago.

There are two challenges in bringing deep learning to wide adoption in multimedia applications (and other applications for that matter).  The first challenge is {\em usability}, namely the implementation of different models and training algorithms must be done by non-experts with little effort. The user must be able to choose among many existing deep learning models, as different multimedia applications may benefit from different models.  For instance, the deep convolution neural network (DCNN) is suitable for image classification~\cite{DBLP:conf/nips/KrizhevskySH12}, recurrent neural network (RNN) for language modelling~\cite{DBLP:conf/icassp/MikolovKBCK11}, and deep auto-encoders for multi-modal data analysis~\cite{DBLP:journals/pvldb/WangOYZZ14,DBLP:conf/mm/FengWL14,DBLP:conf/mm/ZhangYLYC14}. Furthermore, the user must not be required to implement most of these models and training algorithms from scratch, for they are too complex and costly.  An example of complex models is the GoogleLeNet~\cite{DBLP:journals/corr/SzegedyLJSRAEVR14} which comprises 22 layers of 10 different types. Training algorithms are intricate in details. For instance the Back-Propagation~\cite{DBLP:conf/nips/LeCunBOM96} algorithm is notoriously difficult to debug.

The second challenge is {\em scalability}, that is the deep learning system must be able to provision for a huge demand of computing resources for training large models with massive datasets.  As larger training datasets and bigger models are being used to improve  accuracy~\cite{DBLP:journals/corr/abs-1003-0358,DBLP:conf/icml/LeRMDCCDN12,DBLP:journals/corr/SzegedyLJSRAEVR14}, memory requirement for training the model may easily exceed the capacity of a single CPU or GPU. In addition, the computational cost of training may be too high for a single commodity server, which results in unreasonably long training time. For instance, it takes 10 days~\cite{DBLP:journals/corr/YadanATR13,DBLP:journals/corr/PaineJYLH13} to train the DCNN~\cite{DBLP:conf/nips/KrizhevskySH12} with 1.2 million training images and 60 million parameters using one GPU~\footnote{According to the authors, with 2 GPUs, the training still took about 6 days.}.

Addressing both usability and scalability challenges requires a distributed training platform that supports various deep learning models, that comes with an intuitive programming model (similar to MapReduce~\cite{DBLP:conf/osdi/DeanG04}, Spark~\cite{DBLP:conf/nsdi/ZahariaCDDMMFSS12} and epiC~\cite{DBLP:journals/pvldb/Jiang0OTW14} in spirit), and that is scalable. Popular deep learning systems, including Caffe~\cite{jia2014caffe}, Torch~\cite{collobert:2011c} and Theano~\cite{Bastien-Theano-2012}, address the first challenge but fall short at the second challenge (they are not designed for distributed training). Similarly, Google's deep learning platform, called TensorFlow~\cite{tensorflow2015-whitepaper}, is designed to be flexible and easy to use, but its scalability remains unknown (TensorFlow only provides single node version for the time being).  There are several systems supporting distributed training~\cite{DBLP:journals/corr/PaineJYLH13,DBLP:journals/corr/YadanATR13,DBLP:journals/corr/Krizhevsky14}, but they are model specific and do not generalize well to other models. General distributed platforms such as MapReduce and Spark achieve good scalability, but they are designed for general data processing.  As a result, they lack both the programming model and system optimization specific to deep learning, hindering the overall usability and scalability. Recently, there are several specialized distributed platforms~\cite{DBLP:conf/nips/DeanCMCDLMRSTYN12,DBLP:conf/icml/CoatesHWWCN13,186212} that exploit deep learning specific optimization and hence are able to achieve high training throughput. However, they forgo usability issues: the platforms are closed-source and no details of their programming models are given, rendering them unusable by multimedia users.

In this paper, we present our effort in bringing deep learning to the masses. In particular, we extend our previous work~\cite{singa-oss,singa-mm} on distributed training of deep learning models. In \cite{singa-mm}, we designed and implemented an open source distributed deep learning platform, called SINGA\footnote{\url{http://www.comp.nus.edu.sg/~dbsystem/singa/}}, which tackles both usability and scalability challenges at the same time. In this paper, we will introduce optimization techniques and GPU support for SINGA. SINGA provides a simple, intuitive programming model which makes it accessible even to non-experts. SINGA's simplicity is driven by the observation that both the structures and training algorithms of deep learning models can be expressed using a simple abstraction: the neuron layer (or layer). In SINGA, the user defines and connects layers to form the neural network model, and the runtime transparently manages other issues pertaining to the distributed training such as partitioning, synchronization and communication. Particularly, the neural network is represented as a dataflow computation graph with each layer being a node. During distributed training, the graph is partitioned and each sub-graph can be trained on CPUs or on GPUs. SINGA's scalability comes from its flexible system architecture and specific optimization. Both synchronous and asynchronous training frameworks are supported with a range of built-in partitioning strategies, which enables users to readily explore and find an optimal training configuration. Optimization techniques, including minimizing data transferring and overlapping computation and communication, are implemented to reduce the communication overhead from distributed training.


In summary, this paper makes the following contributions:
\begin{enumerate}
	
	\item We present a distributed platform called SINGA which is designed to train deep learning models for multimedia and other applications. SINGA offers a simple and intuitive programming model based on the layer abstraction.
	
	
	\item We describe SINGA's distributed architecture and optimization for reducing the communication overhead in distributed training. 
	
	\item We demonstrate SINGA's usability by describing the implementation of three multimedia applications: multi-modal retrieval, dimensionality reduction and sequence modelling.
	
	\item We evaluate SINGA's performance by comparing it with other open-source systems. The results show that SINGA  is scalable and outperforms other systems in terms of training time.
	
\end{enumerate}

This paper is an extension of our conference paper~\cite{singa-mm}.
In \cite{singa-mm}, we have presented the basic SINGA framework for a homogeneous architecture (where we consider only CPU nodes). In this paper, we extend the framework to a heterogeneous setting that consists of both GPU and CPU processors. Optimization techniques in terms of reducing communication overhead from distributed training are introduced in this paper. Correspondingly, we conducted experiments on GPUs in comparison with existing systems. The rest of this paper is organized as follows. Section~\ref{sec:background} provides the background on training deep learning models and related work. An overview of SINGA as a platform follows in Section~\ref{sec:overview}. The programming model is discussed in Section~\ref{sec:pro-model}. We discuss SINGA architecture and training optimization in Section~\ref{sec:architecture}. The experimental study is presented in Section~\ref{sec:experiment} before we conclude in Section~\ref{sec:conclusion}.

\section{Background}\label{sec:background}

Deep learning is considered as a feature learning technique. A deep learning model typically consists of multiple layers, each associated with a feature transformation function.  After going through all layers, raw input features (e.g., pixels of images) are converted into high-level features that are used for the task of interest, e.g., image classification.

\subsection{Models and Training Algorithms}

\begin{figure}[ht]
	\centering
	\includegraphics[width=.36\textwidth]{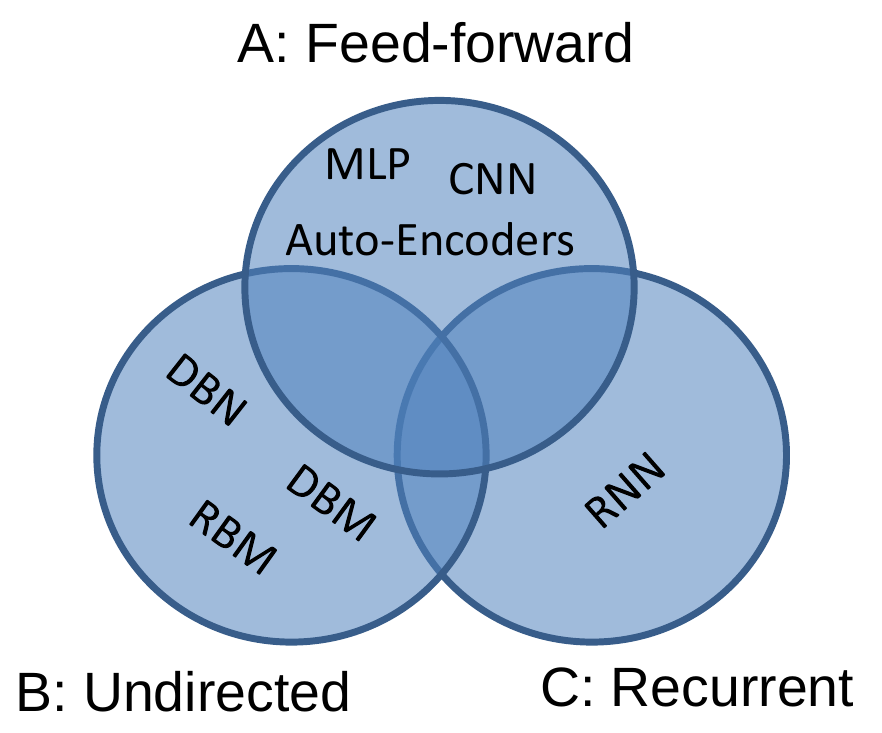}
	\caption{Deep learning model categorization.}
	\label{fig:models}
\end{figure}

We group popular deep learning models into three categories based on the connection types between layers, as shown in Figure~\ref{fig:models}. Category \emph{A} consists of {\em feed-forward models} wherein the layers are directly connected. The extracted features at higher layers are fed into prediction or classification tasks, e.g., image classification~\cite{DBLP:conf/nips/KrizhevskySH12}. Example models in this category include Multi-Layer Perceptron (MLP), Convolution Neural Network (CNN) and Auto-Encoders. Category \emph{B} contains models whose layer connections are undirected. These models are often used to pre-train other models~\cite{HinSal06}, e.g., feed-forward models.  Deep Belief Network (DBN), Deep Boltzmann Machine (DBM) and Restricted Boltzmann Machine (RBM) are examples of such models. Category \emph{C} comprises models that have recurrent connections.  These models are called Recurrent Neutral Networks (RNN). They are widely used for modelling sequential data in which prediction of the next position is affected by previous positions. Language modelling~\cite{DBLP:conf/icassp/MikolovKBCK11} is a popular application of RNN. 

\begin{figure}[ht]
	\centering
	\includegraphics[width=.6\textwidth]{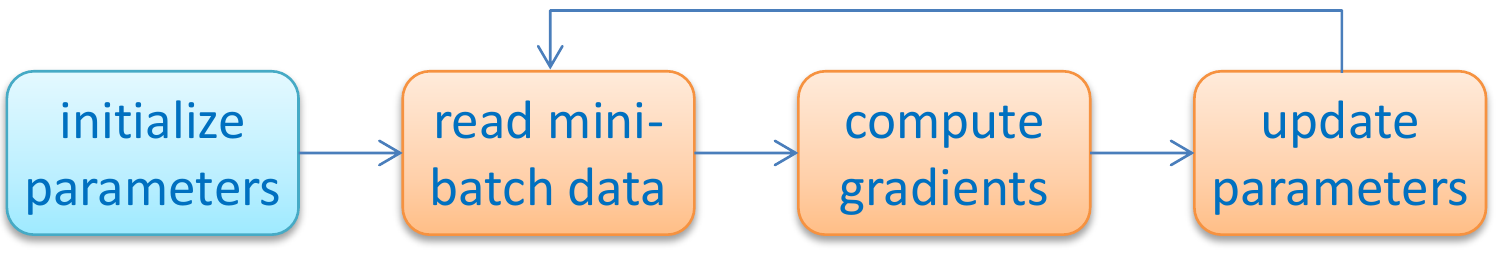}
	\caption{Flow of stochastic gradient descent algorithm.}
	\label{fig:sgd}
\end{figure}

A deep learning model has to be trained to find the optimal parameters for the transformation functions. The training quality is measured by a loss function (e.g., cross-entropy loss) for each specific task. Since the loss functions are usually non-linear and non-convex, it is difficult to get closed-form solutions.  A common approach is to use the Stochastic Gradient Descent (SGD) algorithm shown in Figure~\ref{fig:sgd}.  SGD initializes the parameters with random values, and then iteratively refines them to reduce the loss based on the computed gradients. There are three typical algorithms for gradient computation corresponding to the three model categories above: Back-Propagation (BP), Contrastive Divergence (CD) and Back-Propagation Through Time (BPTT).

\begin{table}[ht]
	\centering
	\caption{Feature comparison with other open source systems.\label{tb:features}}
	\begin{tabular}{|l|l|l|l|l|l|l|l|}
		\hline		
		& SINGA & Tensor- & Caffe & Torch 	&	MxNet &	Theano &		Cuda-\\
		& &Flow &  &  & 	&	&	convnet2\\\hline			
		Feed-forward net&\checkmark&\checkmark&\checkmark&\checkmark&\checkmark&\checkmark&\checkmark\\\hline
		Energy model&\checkmark&\checkmark&$\times$&\checkmark&-&\checkmark&$\times$\\\hline
		RNN&\checkmark&\checkmark&\checkmark&\checkmark&\checkmark&\checkmark&$\times$\\\hline
		\textbf{Data parallelism}&\checkmark&\checkmark&\checkmark&\checkmark&\checkmark&-&\checkmark\\\hline
		\textbf{Model parallelism}&\checkmark&-&-&-&-&-&\checkmark\\\hline
		\textbf{Hybrid parallelism}&\checkmark&-&-&-&-&-&\checkmark\\\hline
		GPU&\checkmark&\checkmark&\checkmark&\checkmark&\checkmark&\checkmark&\checkmark\\\hline
		CPU&\checkmark&\checkmark&\checkmark&\checkmark&\checkmark&\checkmark&$\times$\\\hline
		Python&\checkmark&\checkmark&\checkmark&$\times$&\checkmark&\checkmark&\checkmark\\\hline
		R&working&-&\checkmark&-&\checkmark&$\times$&$\times$\\\hline
		HDFS&\checkmark&-&-&-&\checkmark&-&$\times$\\\hline
		Mesos/YARN&\checkmark&-&-&-&-&-&$\times$\\\hline
		\multicolumn{8}{l}{-: Could support but not implemented yet.}
	\end{tabular}
\end{table}

\subsection{Related Work}\label{sec:related}

Due to its outstanding capabilities in capturing complex regularities of multimedia data (e.g., image and video), deep learning techniques are being adopted by more and more multimedia applications, e.g., image retrieval~\cite{DBLP:conf/mm/WanWHWZZL14}, multi-modal retrieval~\cite{raey,DBLP:journals/pvldb/WangOYZZ14}, sentiment analysis~\cite{DBLP:conf/mm/YouLJY15}, etc. In recent years, we have witnessed fast increase of deep learning models' depth, from tens of layers (e.g., AlexNet~\cite{DBLP:conf/nips/KrizhevskySH12}, VGG~\cite{DBLP:journals/corr/SimonyanZ14a}) to hundreds of layers~\cite{he2015deep}. It has been shown that deeper models work better for the ImageNet challenge task~\cite{DBLP:journals/corr/SzegedyLJSRAEVR14,DBLP:journals/corr/SimonyanZ14a}. Meanwhile, training datasets are also becoming larger, from 60,000 images in the MNIST and Cifar datasets to millions of images in the ImageNet dataset. Complex deep models and massive training datasets require a huge amount of computing resources for training.



Different applications use different deep learning models. It is essential to provide a general deep learning system for non-experts to implement their models without much effort. Recently, some distributed training approaches have been proposed, for examples~\cite{DBLP:journals/corr/PaineJYLH13,DBLP:journals/corr/YadanATR13,DBLP:journals/corr/Krizhevsky14}. They are specifically optimized for training the AlexNet model~\cite{DBLP:conf/nips/KrizhevskySH12}, thus cannot generalize well to other models. Other general distributed deep learning platforms~\cite{DBLP:conf/nips/DeanCMCDLMRSTYN12,DBLP:conf/icml/CoatesHWWCN13,186212} exploit deep learning specific optimization and hence are able to achieve high training throughput. However, they are closed-source and there are no details of the programming model, rendering them unusable to developers. There are also some popular open source systems for training deep learning models on a single node, including TensorFlow~\cite{tensorflow2015-whitepaper}, Caffe~\cite{jia2014caffe}, Torch~\cite{collobert:2011c}, MxNet~\cite{chen2015mxnet}, Theano~\cite{Bastien-Theano-2012} and Cuda-Convnet2\footnote{https://code.google.com/p/cuda-convnet/}. Table~\ref{tb:features} shows the comparison of SINGA and these systems in terms of supported features.

\section{Overview}\label{sec:overview}
\begin{figure}[ht]
	\centering
	\includegraphics[width=.5\textwidth]{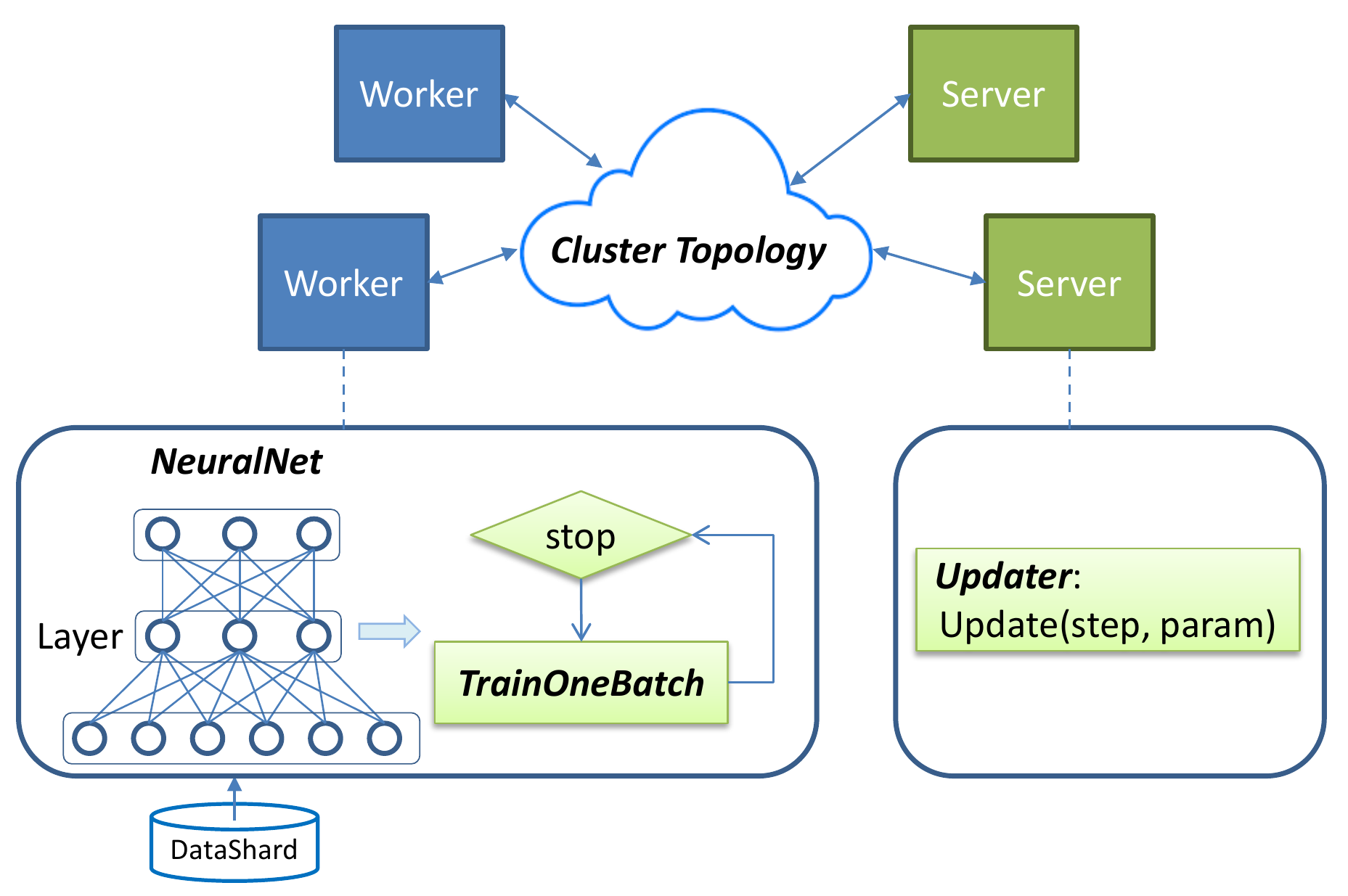}
	\caption{SINGA overview.}
	\label{fig:overview}
\end{figure}

SINGA trains deep learning models using SGD over the worker-server architecture, as shown in Figure~\ref{fig:overview}. Workers compute parameter gradients and servers perform parameter updates.  To start a training job, the user (or programmer) submits a job configuration specifying the following four components:
\begin{itemize}
\item
A \emph{NeuralNet} describing the neural network (or neural net) structure with the detailed layers and their connections. SINGA comes with many built-in layers (Section~\ref{sec:layer}), and users can also implement their own layers.

\item
A \emph{TrainOneBatch} algorithm for training the model. SINGA implements different algorithms (Section~\ref{sec:trainonebatch}) for all three model categories.

\item
An \emph{Updater} defining the protocol for updating parameters at the servers (Section~\ref{sec:updater}).

\item
A \emph{Cluster Topology} specifying the distributed architecture of workers and servers. SINGA's architecture is flexible and can support both synchronous and asynchronous training (Section~\ref{sec:architecture}).
\end{itemize}

Given a job configuration, SINGA distributes the training tasks over the cluster and coordinates the
training. In each iteration, every worker calls \emph{TrainOneBatch} function to compute parameter gradients.
\emph{TrainOneBatch} takes a \emph{NeuralNet} object representing the neural net, and it visits (part of) the
model layers in an order specific to the model category. The computed gradients are sent to the corresponding
servers for updating. Workers then fetch the updated parameters at the next iteration.

\section{Programming Model}\label{sec:pro-model}

\begin{figure}[ht]
	\subfigure[Sample MLP.]{%
		\includegraphics[width=0.3\textwidth]{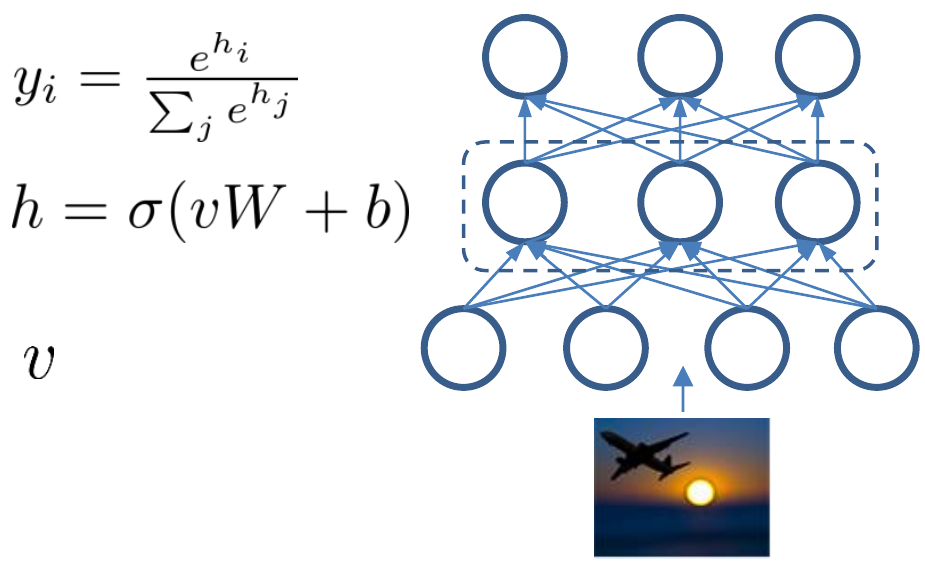}\label{fig:mlp-sample}}
	\subfigure[NeuralNet configuration.]{ %
		\includegraphics[width=0.3\textwidth]{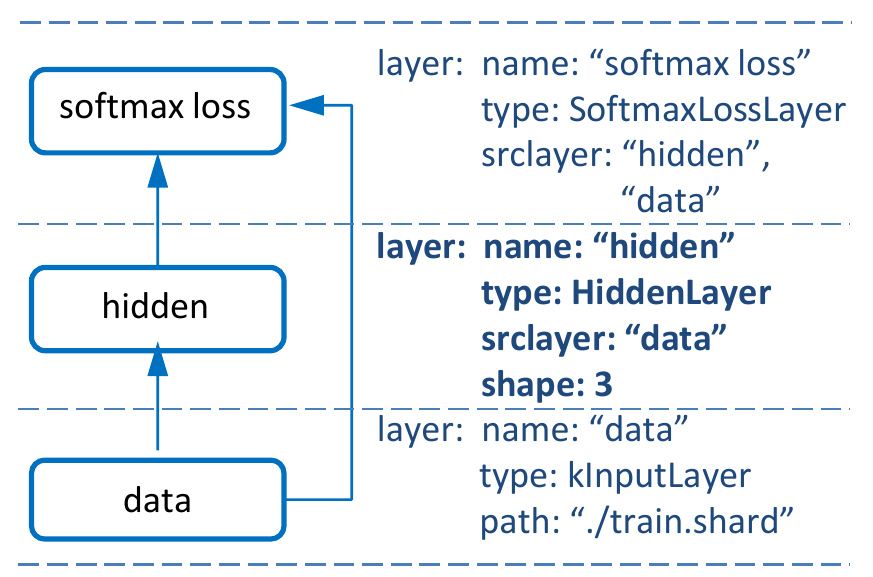}\label{fig:mlp-config}}
	\subfigure[Hidden layer implementation.]{ %
		\includegraphics[width=.35\textwidth]{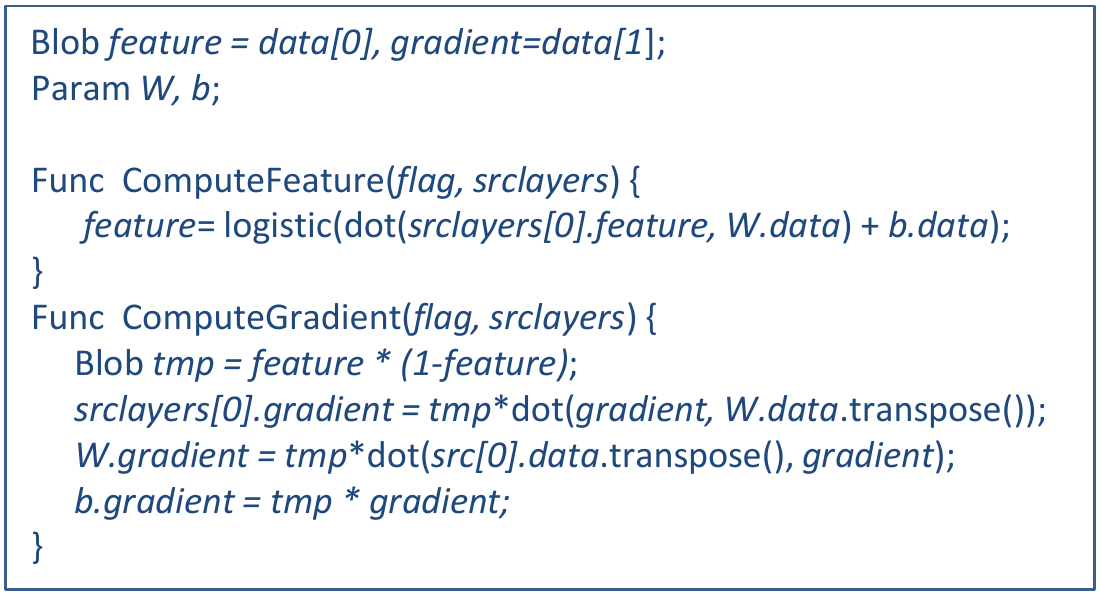}\label{fig:mlp-layer}}
	\caption{Running example using an MLP.}
	\label{fig:mlp}
\end{figure}

This section describes SINGA's programming model, particularly the main components of a SINGA job. We use the MLP model for image classification (Figure~\ref{fig:mlp-sample}) as a running example. The model consists of an input layer, a hidden feature transformation layer and a Softmax output layer.

\subsection{Programming Abstractions} \label{sec:pro-abstraction}

\subsubsection{NeuralNet}\label{sec:net}

\emph{NeuralNet} represents a neural net instance in SINGA. It comprises a set of unidirectionally connected layers. Properties and connections of layers are specified by users. The {\em NeuralNet} object is passed as an argument to the \emph{TrainOneBatch} function.

\begin{figure}[ht]
	\centering
	\subfigure[\label{fig:convert-rbm}Convert connections in RBM.]{%
		\includegraphics[width=.3\textwidth]{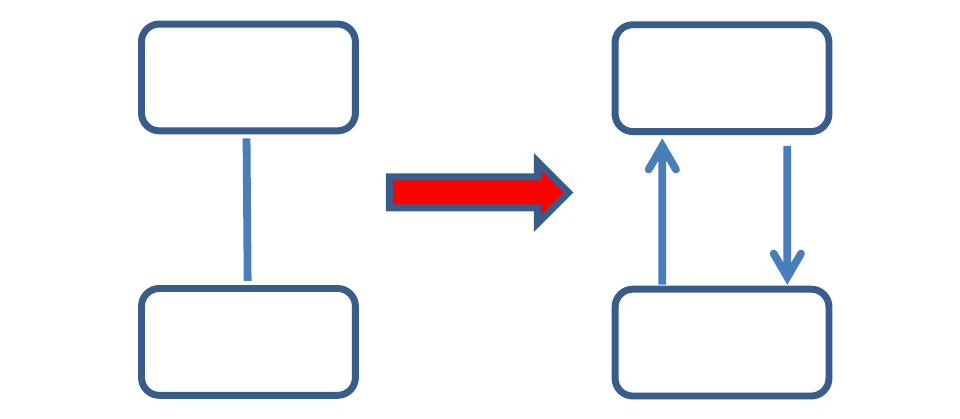}}
	\qquad
	\subfigure[\label{fig:unroll-rnn}Unroll RNN.]{ %
		\includegraphics[width=.5\textwidth]{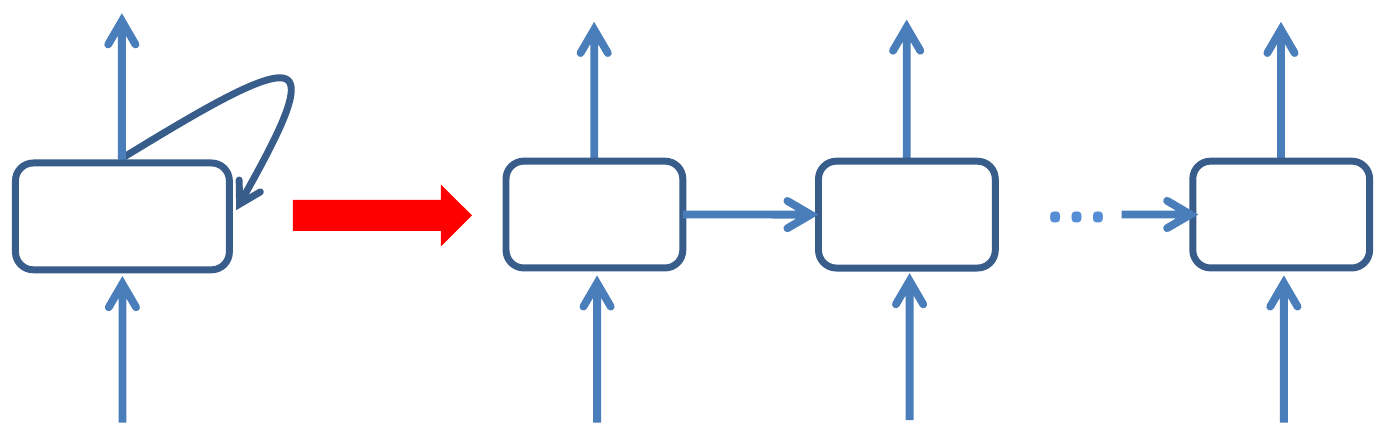}}
	\caption{Unify neural net connections.}
	\label{fig:rbm-rnn}
\end{figure}

Layer connections in {\em NeuralNet} are not designed explicitly; instead each layer records its own source layers as specified by users (Figure~\ref{fig:mlp-config}). Although different model categories have different types of layer connections, they can be unified using directed edges as follows. For feed-forward models, nothing needs to be done as their connections are already directed.  For undirected models, users need to replace each edge with two directed edges, as shown in Figure~\ref{fig:convert-rbm}. For recurrent models, users can unroll a recurrent layer into directed-connecting sub-layers, as shown in Figure~\ref{fig:unroll-rnn}.

\subsubsection{Layer}\label{sec:layer}

\emph{Layer} is a core abstraction in SINGA. Different layer implementations perform different feature transformations to extract high-level features. In every SGD iteration, all layers in the \emph{NeuralNet} are visited by the \emph{TrainOneBatch} function during the process of computing parameter gradients. From the dataflow perspective, we can regard the neural net as a graph where each layer is a node. The training procedure passes data along the connections of layers and invokes functions of layers. Distributed training can be easily conducted by assigning sub-graphs to workers.

\begin{figure}[ht]
	\centering
  \includegraphics[width=.5\textwidth]{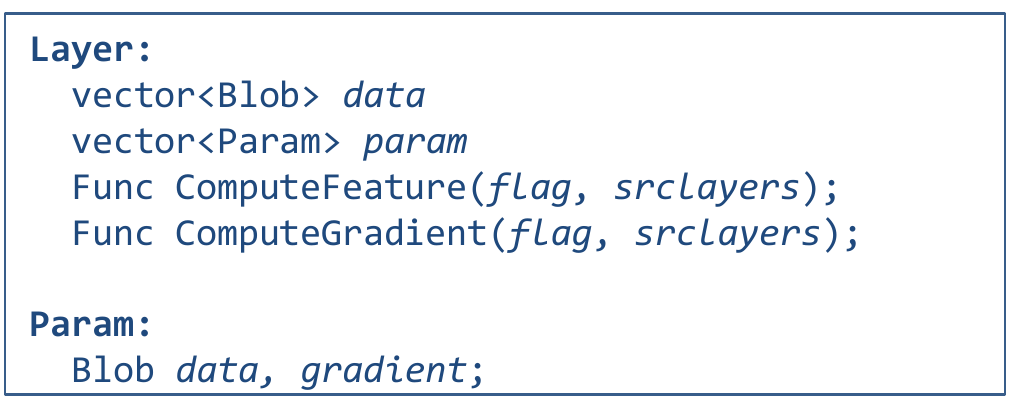}
 	\caption{Layer abstraction.}
	\label{fig:layer}
\end{figure}

Figure~\ref{fig:layer} shows the definition of a base layer.  The \emph{data} field records data (blob) associated with a layer. Some layers may require parameters (e.g., a weight matrix) for their feature transformation functions. In this case, these parameters are represented by \emph{Param} objects, each with a \emph{data} field for the parameter values and a \emph{gradient} field for the gradients. The \emph{ComputeFeature} function evaluates the feature blob by transforming features from the source layers.  The \emph{ComputeGradient} function computes the gradients associated with this layer. These two functions are invoked by the \emph{TrainOneBatch} function during training (Section~\ref{sec:trainonebatch}).

SINGA provides a variety of built-in layers to help users build their models. Table~\ref{tb:layers} lists the layer categories in SINGA. For example, the data layer loads a mini-batch of records via the \emph{ComputeFeature} function in each iteration.  Users can also define their own layers for their specific requirements.  Figure~\ref{fig:mlp-layer} shows an example of implementing the hidden layer $h$ in the MLP.  In this example, beside  feature blobs there are gradient blobs storing the gradients of the loss with respect to the feature blobs. There are two \emph{Param} objects: the weight matrix $W$ and the bias vector $b$.  The \emph{ComputeFeature} function rotates (multiply $W$), shifts (plus $b$) the input features and then applies non-linear (logistic) transformations.  The \emph{ComputeGradient} function computes the layer's parameter gradients, as well as the source layer's gradients that will be used for evaluating the source layer's parameter gradients.

\begin{table}[ht]
	\centering	
	\caption{Layer categories.\label{tb:layers}}
	\begin{tabular}{|l|l|}
		\hline
		\textbf{Category}&{\bf Description}\\\hline
		Input layers & Load records from file, database or HDFS. \\
		\hline
		Output layers & Dump records to file, database or HDFS. \\
		\hline
		Neuron layers & Feature transformation, e.g., convolution.\\
		\hline
		Loss layers & Compute objective loss, e.g., cross-entropy loss.\\
		\hline
		Connection layers & Connect layers when neural net is partitioned.\\\hline
	\end{tabular}
\end{table}

\subsubsection{TrainOneBatch}\label{sec:trainonebatch}

The \emph{TrainOneBatch} function determines the sequence of invoking
\emph{ComputeFeature} and \emph{ComputeGradient} functions in all layers during
each SGD iteration.  SINGA implements two \emph{TrainOneBatch} algorithms for the three model categories. For feed-forward and recurrent models, the BP algorithm is provided. For undirected modes (e.g., RBM), the CD algorithm is provided. Users simply select the corresponding algorithm in the job configuration.  Should there be specific requirements for the training workflow, users can define their own \emph{TrainOneBatch} function following a template shown in Algorithm~\ref{alg:bp}.  Algorithm~\ref{alg:bp} implements the BP algorithm which takes a \emph{NeuralNet} object as input. The first loop visits each layer and computes their features, and the second loop visits each layer in the reverse order and computes parameter gradients. More details on applying BP for RNN models (i.e., BPTT), and the CD algorithm.

\begin{algorithm}[ht]
	\caption{BPTrainOneBatch}
	\label{alg:bp}
	\KwIn{net}
	\ForEach{layer in net.layers}{
		Collect(layer.params()) \tcp{receive parameters}
		layer.ComputeFeature() \tcp{forward prop}
	}
	\ForEach{ layer in reverse(net.layers)}{
		layer.ComputeGradient()\tcp{backward prop}
		Update(layer.params())\tcp{send gradients}
	}
\end{algorithm}

\subsubsection{Updater}
\label{sec:updater}
Once the parameter gradients are computed, workers send these values to servers to update the parameters.  SINGA implements several parameter updating protocols, such as AdaGrad\cite{DBLP:journals/jmlr/DuchiHS11}. Users can also define their own updating protocols by overriding the \emph{Update} function.

\subsection{Multimedia Applications}
\label{sec:app}

This section demonstrates the use of SINGA for multimedia applications. We discuss the training of three deep
learning models for three different applications: a multi-modal deep neural network (MDNN) for multi-modal retrieval, a RBM
for dimensionality reduction, and a RNN for sequence modelling.

\subsubsection{MDNN for Multi-modal Retrieval}
\label{sec:feedforward}
\begin{figure}[ht]
	\centering
	\includegraphics[width=.56\textwidth]{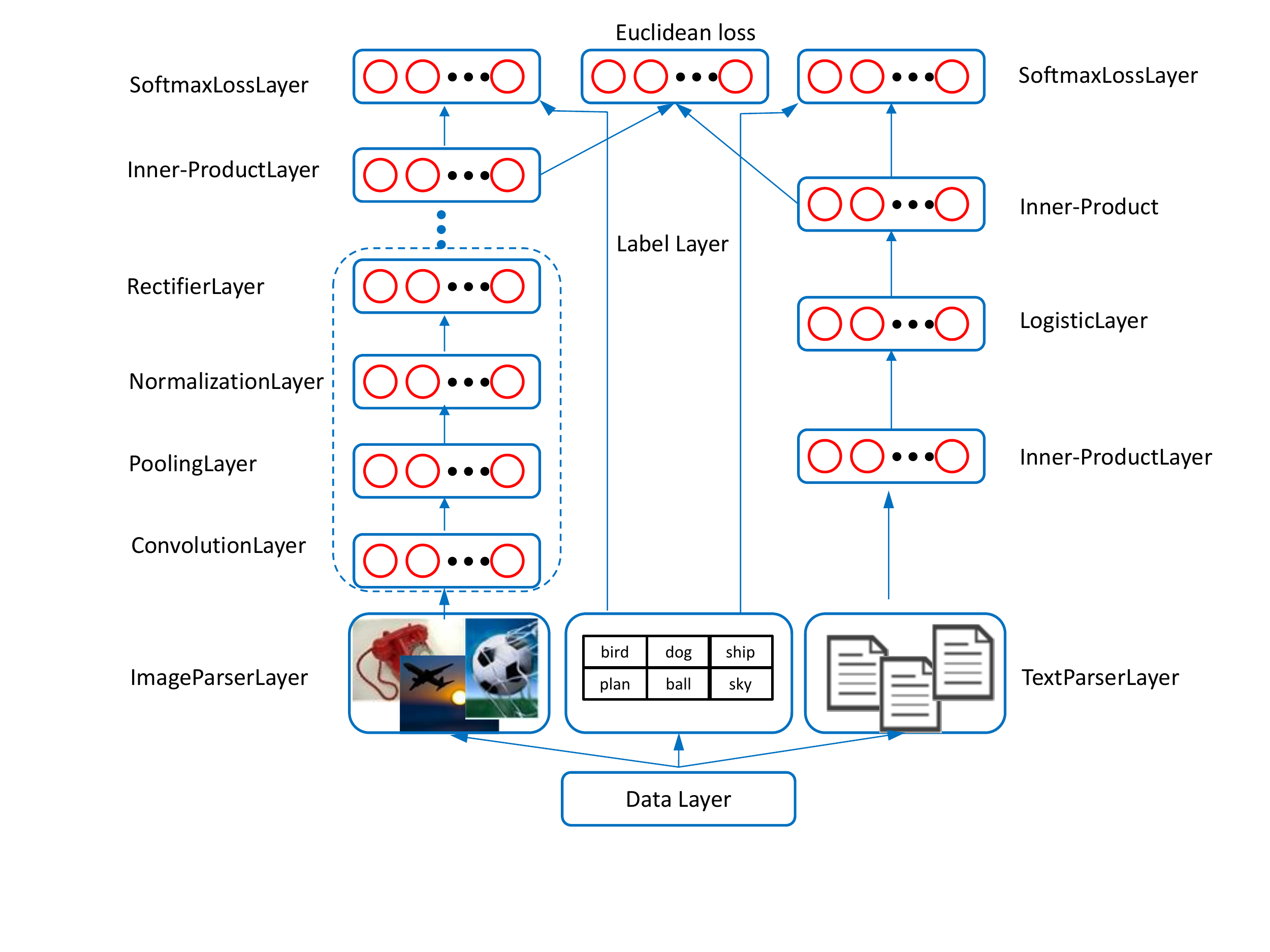}
	\caption{Structure of MDNN.}
	\label{fig:mdnn-net}
\end{figure}

Feed-forward models such as CNN and MLP are widely used
to learn high-level features in multimedia applications, especially for image
classification~\cite{DBLP:conf/nips/KrizhevskySH12}. Here, we demonstrate the training of the
MDNN~\cite{raey}, which combines a CNN and a MLP. MDNN is used for extracting features
for the multi-modal retrieval task~\cite{DBLP:journals/pvldb/WangOYZZ14,DBLP:conf/mm/FengWL14,DBLP:conf/mm/ShenOT00} that searches objects from different modalities. In MDNN, the
CNN~\cite{DBLP:conf/nips/KrizhevskySH12} is used to extract image features, and the MLP is used to extract text
features. The training objective is to minimize a weighted sum of: (1) the error of predicting the labels of image and text documents using extracted features;
and (2) the distance between features of relevant image and text objects. As a result, the learned features of
semantically relevant objects from different modalities are similar. After training, multi-modal retrieval is
conducted using the learned features.

Figure~\ref{fig:mdnn-net} depicts neural net of MDNN model in SINGA. We can see that there are two parallel
paths: one for text modality and the other for image modality. The data layer reads in records of
semantically relevant image-text pairs. The image layer, text layer and label layer then parse the visual feature,
text feature (e.g., tags of the image) and labels respectively from the records. The image path consists of
layers from DCNN~\cite{DBLP:conf/nips/KrizhevskySH12}, e.g., the convolution layer and pooling layer. The
text path includes an inner-product (or fully connected) layer, a logistic layer and a loss layer. The
Euclidean loss layer measures the distance of the feature vectors extracted from these two paths. All
except the parser layers, which are application specific, are SINGA's built-in layers. Since this model is a feed-forward model, the BP algorithm is selected for the \emph{TrainOneBatch} function.

\subsubsection{RBM for Dimensionality Reduction}
\label{sec:undirected}

RBM is often employed to pre-train parameters for other models. In this example application, we use RBM to pre-train a deep auto-encoder~\cite{HinSal06} for  dimensionality reduction. Multimedia applications typically operate with high-dimensional feature vectors, which demands large computing resources. Dimensionality reduction techniques, such as Principal Component Analysis (PCA), are commonly applied in the pre-processing step.
Deep auto-encoder is reported~\cite{HinSal06} to have better performance than PCA.

\begin{figure}[ht]
	\centering
	\includegraphics[width=.65\textwidth]{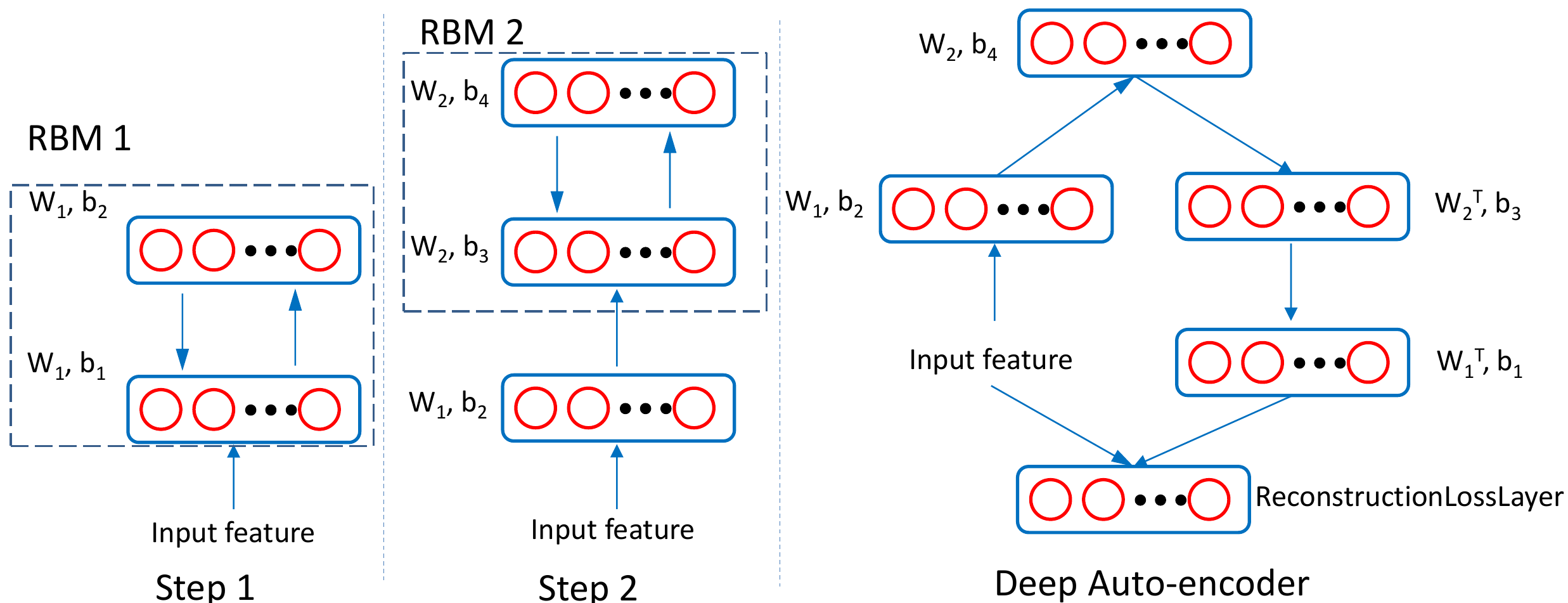}
	\caption{Structure of RBM and deep auto-encoder.}
	\label{fig:rbm-net}
\end{figure}

Generally, the deep auto-encoder is trained to reconstruct the input feature using the feature of the top
layer. Hinton et al.~\cite{HinSal06} used RBM to pre-train the parameters for each layer, and fine-tuned them to minimize the
reconstruction error. Figure~\ref{fig:rbm-net} shows the model structure (with parser layer and data layer
omitted) in SINGA. The parameters trained from the first RBM
(RBM 1) in step 1 are ported (through checkpoint) into step 2 wherein the extracted features are used to
train the next model (RBM 2).  Once pre-training is finished, the deep auto-encoder is unfolded for
fine-tuning.  SINGA applies the contrastive divergence (CD) algorithm for training RBM and back-propagation (BP) algorithm for fine-tuning the deep auto-encoder.

\subsubsection{RNN for Sequence Modelling}
\label{sec:recurrent}

Recurrent neural networks (RNN) are widely used for modelling sequential data, e.g., natural language sentences. We use SINGA to train a Char-RNN model \footnote{https://github.com/karpathy/char-rnn} over Linux kernel source code, with each character as an input unit. The model predicts the next character given the current character.

\begin{figure}[ht]
	\centering
	\includegraphics[width=.8\textwidth]{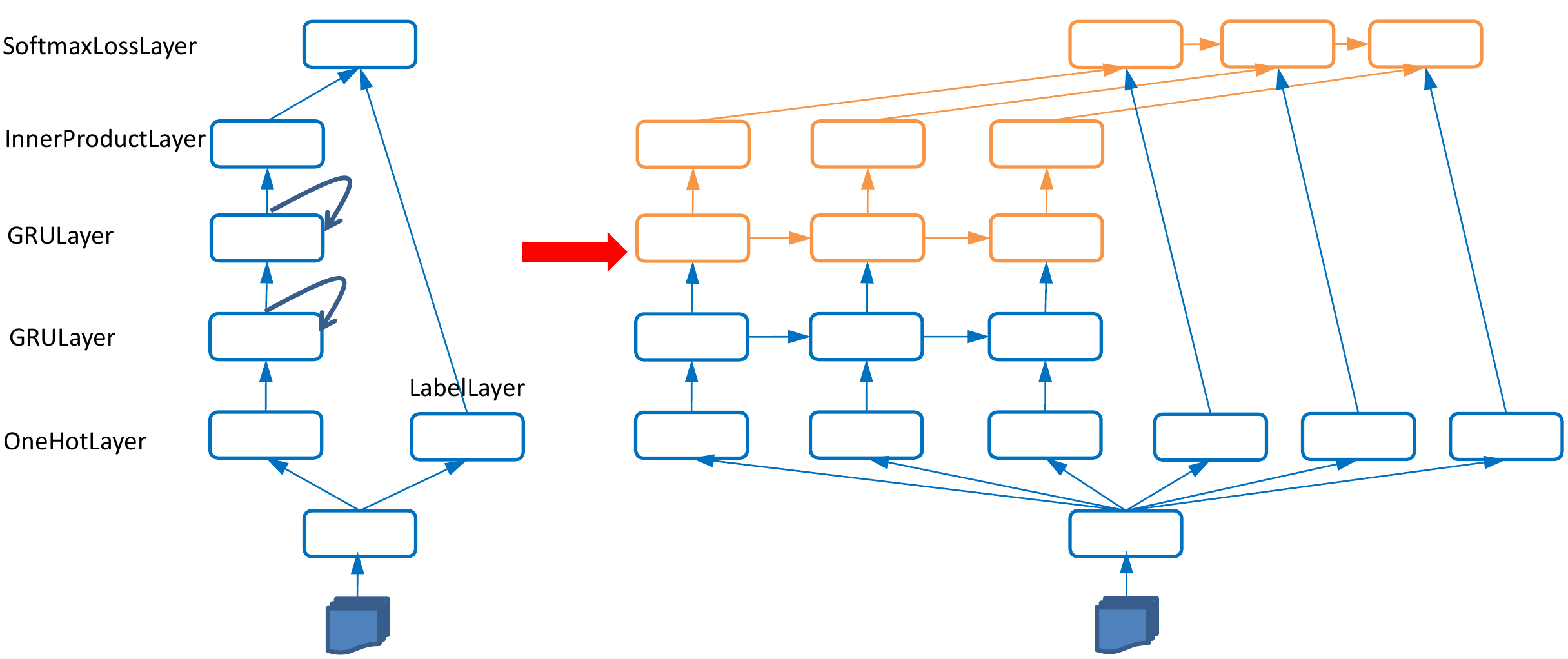}
	\caption{Structure of 2-stacked Char-RNN (left before unrolling; right after unrolling).}
	\label{fig:rnn-net}
\end{figure}

Figure~\ref{fig:rnn-net} illustrates the net structure of the Char-RNN model. The input layer buffers all training data (the Linux kernel code is about 6MB). In each iteration, it reads $unroll\_len+1$ ($unroll\_len$ is specified by users) successive characters, e.g., ``int a;'' and passes the first $unroll\_len$ characters to OneHotLayers (one per layer). Each OneHotLayer converts its character into a one-hot vector representation. The input layer passes the last $unroll\_len$ characters as labels to the RNNLabelLayer (the label of the $i^{th}$ character is the $(i+1)^{th}$ character, i.e., the objective is to predict the next character). Each GRULayer receives a one-hot vector and the hidden feature vector from its precedent layer. After some feature transformations, its own feature vector is passed to another stack of GRULayer and its successive GRULayer. The InnerProductLayers transform the output from the GRULayers in the second stack and feed them into the SoftmaxLossLayer. The $i^{th}$ SoftmaxLossLayer measures the cross-entropy loss for predicting the $i^{th}$ character. The model is configured similarly as for feed-forward models except the training algorithm is  BPTT, and unrolling length and connection types are specified for recurrent layers. Different colors are used for illustrating the neural net partitioning which will be discussed in Section~\ref{sec:partition}.

\section{Distributed Training}\label{sec:architecture}
In this section, we introduce SINGA's architecture, and discuss how it supports a variety of distributed training frameworks.

\subsection{System Architecture}\label{subsec:arch}
\begin{figure}
	\centering
	\includegraphics[width=.56\textwidth]{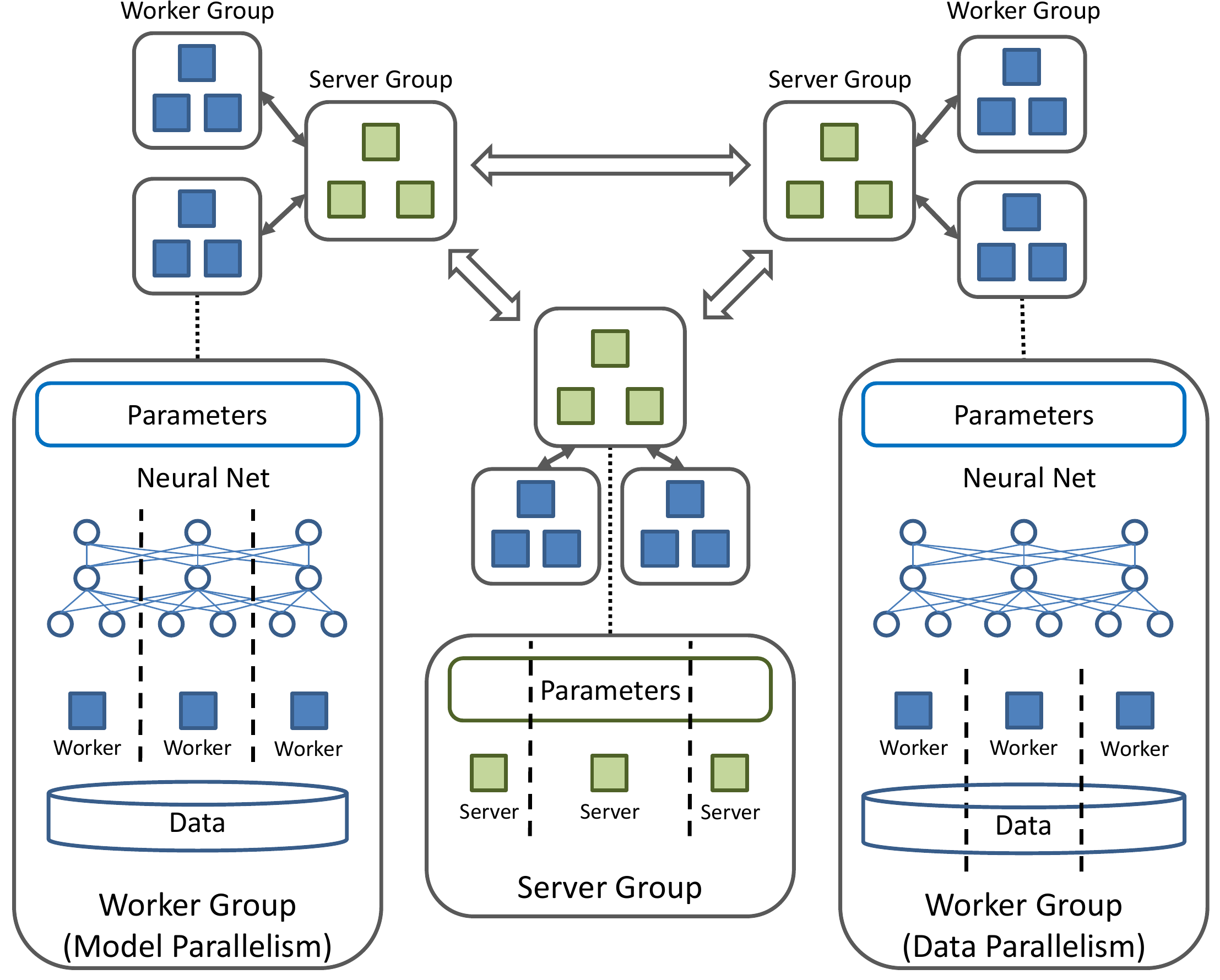}
	\caption{Logical architecture of SINGA.}
	\label{fig:logical-arch}
\end{figure}

Figure~\ref{fig:logical-arch} shows the logical architecture, which consists of multiple server groups
and worker groups, and each worker group communicates with only one server group.  Each server group
maintains a complete replica of the model parameters, and is responsible for handling requests (e.g., get or update
parameters) from worker groups.  Neighboring server groups synchronize their parameters periodically.
Typically, a server group contains a number of servers, and each server manages a partition of the model
parameters.  Each worker group trains a complete model replica against a partition of the training
dataset (i.e. data parallelism), and is responsible for computing parameter gradients. All worker groups run
and communicate with the corresponding server groups asynchronously.  However, inside each worker group, the workers
 compute parameter updates synchronously for the model replica.  There are two strategies to distribute the
training workload among workers within a group: by model or by data. More specifically, each worker can compute a subset
of parameters against all data partitioned to the group (i.e., model parallelism), or all parameters against a subset of data
(i.e., data parallelism). SINGA also supports hybrid parallelism (Section~\ref{sec:partition}).

In SINGA, servers and workers are execution units running in separate threads. If GPU devices are available, SINGA automatically assigns $g$ GPU devices ($g$ is user specified) to the first $g$ workers on each node. A GPU worker executes the layer functions on GPU if they are implemented using GPU API (e.g., CUDA). Otherwise, the layer functions execute on CPU. SINGA provides several linear algebra functions for users to implement their own layer functions. These linear algebra functions have both GPU and CPU implementation and they determine the running device of the calling thread automatically. In this way, we keep the implementation transparent to users. Workers and servers communicate through message passing. Every process runs the main thread as a stub that aggregates local messages and forwards them to corresponding (remote) receivers. 



\subsection{Training Frameworks}\label{subsec:frame}
In SINGA, worker groups run asynchronously and workers within one group run synchronously. Users can leverage this general design to run both synchronous and asynchronous training frameworks. Specifically, users control the training framework by configuring the cluster topology, i.e., the number of worker (resp. server) groups and worker (resp. server) group size. In the following, we will discuss how to realize popular distributed training frameworks in SINGA, including Sandblaster and Downpour from Google's DistBelief system~\cite{DBLP:conf/nips/DeanCMCDLMRSTYN12}, AllReduce from Baidu's DeepImage system~\cite{DBLP:journals/corr/WuYSDS15} and distributed Hogwild from Caffe~\cite{jia2014caffe}. 

\begin{figure}[h]
	\centering
	\includegraphics[width=.86\textwidth]{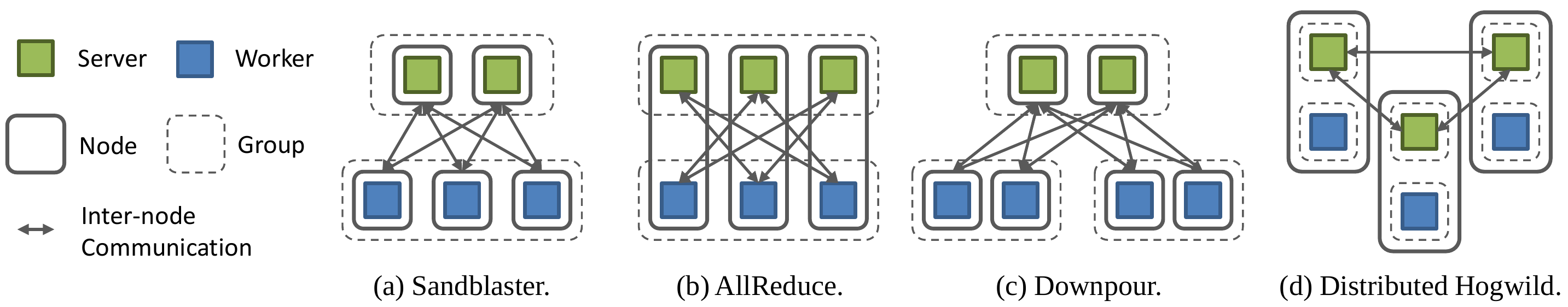}
	\caption{Training frameworks in SINGA.\label{fig:frameworks}}
\end{figure}

\subsubsection{Synchronous Training}

A synchronous framework is realized by configuring the cluster topology with only one worker group and one
server group. The training convergence rate is the same as that on a single node.

Figure~\ref{fig:frameworks}a shows the Sandblaster framework implemented in SINGA. A single server group is
configured to handle requests from workers.  A worker operates on its partition of the model, and
only communicates with servers handling the related parameters.  Figure~\ref{fig:frameworks}b shows the
AllReduce framework in SINGA, in which we bind each worker with a server on the same node, so that each node
is responsible for maintaining a partition of parameters and collecting updates from all other
nodes.

Synchronous training is typically limited to a small or medium size cluster
, e.g. fewer than 100 nodes. When the cluster size is large,
the synchronization delay is likely to be larger than the computation time. Consequently, the training cannot scale well.

\subsubsection{Asynchronous Training}

An asynchronous framework is implemented by configuring the cluster topology with more than one worker groups.
The training convergence is likely to be different from single-node training, because multiple worker groups
are working on different versions of the parameters~\cite{DBLP:dblp_journals/corr/ZhangR14}.

Figure~\ref{fig:frameworks}c shows the Downpour~\cite{DBLP:conf/nips/DeanCMCDLMRSTYN12} framework implemented
in SINGA. Similar to the synchronous Sandblaster, all workers send requests to a global server group.
We divide workers into several  groups, each running independently and working on parameters from the last \emph{update} response.  Figure~\ref{fig:frameworks}d shows the distributed Hogwild framework, in which each node
contains a complete server group and a complete worker group. Parameter updates are done locally, so that
communication cost during each training step is minimized. However, the server group must periodically synchronize with neighboring groups
to improve the training convergence. The topology (connections) of server groups can be customized (the default topology is all-to-all connection).

Asynchronous training can improve the convergence rate to some degree. But the
improvement typically diminishes when there are more model replicas. A
more scalable training framework should combine both the synchronous and
asynchronous training. In SINGA, users can run a hybrid training framework by launching
multiple worker groups that run asynchronously to improve the convergence rate.
Within each worker group, multiple workers run synchronously to
accelerate one training iteration. Given a fixed budget (e.g., number of nodes
in a cluster), there are opportunities to find one optimal hybrid training framework that trades off between the convergence rate and efficiency in order to achieve the minimal training time.

\subsection{Neural Network Partitioning}
\label{sec:partition}
In this section, we describe how SINGA partitions the neural net to support data parallelism, model parallelism, and
hybrid parallelism within one worker group.

\begin{figure}[ht]
	\centering
	\includegraphics[width=.6\textwidth]{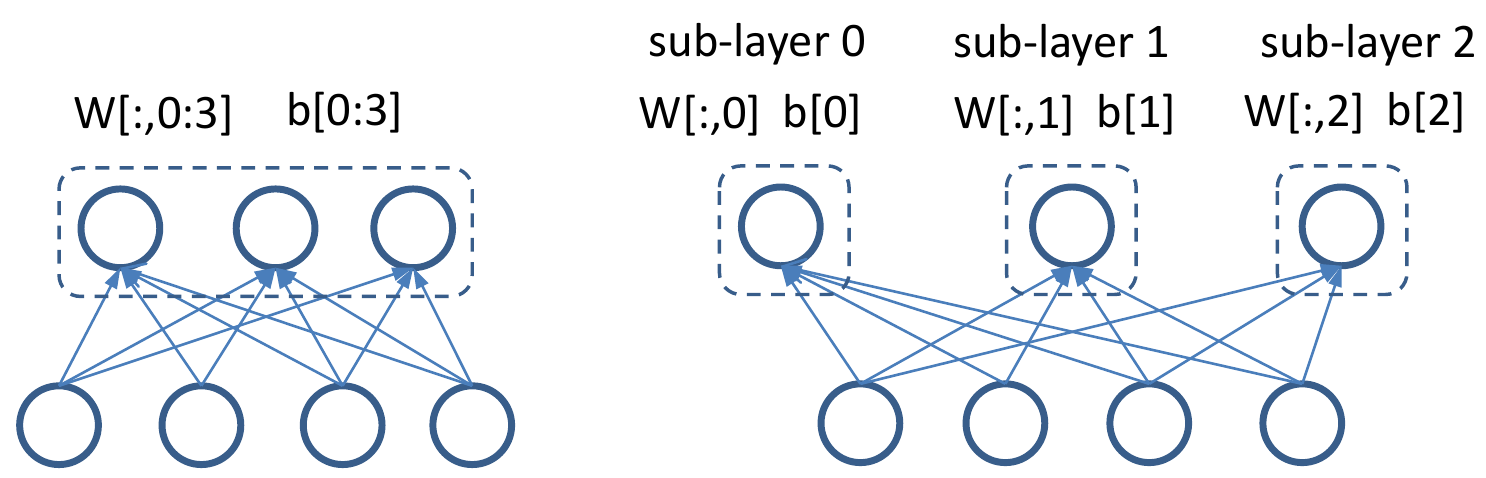}
	\caption{Partition the hidden layer in Figure~\ref{fig:mlp-sample}.}
	\label{fig:partition-layer}
\end{figure}

SINGA partitions a neural net at the granularity of layer. Every layer's feature blob is considered a matrix whose rows are
feature vectors. Thus, the layer can be split on two dimensions. Partitioning on dimension 0 (also called batch dimension) slices
the feature matrix by row. For instance, if the mini-batch size is 256 and the layer is partitioned into 2 sub-layers,
each sub-layer would have 128 feature vectors in its feature blob. Partitioning on this dimension has no effect on the parameters, as every \emph{Param} object is replicated in the sub-layers. Partitioning on dimension 1 (also called feature dimension) slices the
feature matrix by column. For example, suppose the original feature vector has 50 units, after partitioning into 2 sub-layers, each sub-layer would have 25 units. This partitioning splits \emph{Param} objects, as shown in Figure~\ref{fig:partition-layer}. Both the bias vector and weight matrix
are partitioned into two sub-layers (workers).

Network partitioning is conducted while creating the \emph{NeuralNet} instance. SINGA extends a layer into multiple sub-layers. Each sub-layer is assigned a location ID, based on which it is dispatched to the corresponding worker. Advanced users can
also directly specify the location ID for each layer to control the placement of layers onto workers. For the MDNN model in
Figure~\ref{fig:mdnn-net}, users can configure the layers in the image path with location ID 0 and the layers in the text
path with location ID 1, making the two paths run in parallel. Similarly, for the Char-RNN model shown in Figure~\ref{fig:rnn-net}, we can place the layers of different colors onto different workers. Connection layers will be automatically added to connect
the sub-layers. For instance, if two connected sub-layers are located at two different workers, then a pair of
bridge layers is inserted to transfer the feature (and gradient) blob between them. When two layers are partitioned
on different dimensions, a concatenation layer which concatenates feature rows (or columns) and a slice layer
which slices feature rows (or columns) are inserted. Connection layers help make the network
communication and synchronization transparent to the users.

When every worker computes the gradients of the entire model parameters, we refer to this process as data parallelism. 
When different workers compute the gradients of different parameters, we call this process model parallelism. In particular,
partitioning on dimension 0 of each layer results in data parallelism, while partitioning on dimension 1 results in model parallelism. Moreover, SINGA 
supports hybrid parallelism wherein some workers compute the gradients of the same subset of model parameters while
other workers compute on different model parameters. For example, to implement the hybrid parallelism in
\cite{DBLP:journals/corr/Krizhevsky14} for the CNN model, we set \emph{partition\_dim = 0} for  lower layers and
\emph{partition\_dim = 1} for higher layers. The following list summarizes the partitioning strategies, their trade-off is analyzed in Section~\ref{sec:opt}.

\begin{enumerate}
	\item Partitioning all layers into different subsets $\rightarrow$ model parallelism.
	\item Partitioning each singe layer into sub-layers on batch dimension $\rightarrow$ data parallelism.
	\item Partitioning each singe layer into sub-layers on feature dimension $\rightarrow$ model parallelism.
	\item Hybrid partitioning of strategy 1, 2 and 3 $\rightarrow$ hybrid parallelism.
\end{enumerate}


\subsection{Optimization}\label{sec:opt}
Distributed training (i.e, partitioning the neural net and running workers over different layer partitions) increases the computation power, i.e., FLOPS. However, it introduces overhead in terms of communication and synchronization. Suppose we have a homogeneous computation environment, that is, all workers run at the same speed and get the same workload (e.g., same number of training samples and same size of feature vectors). In this case, we can ignore the synchronization overhead and analyze only the communication cost. The communication cost is mainly attributed to the data transferred through PCIe over multiple GPUs in a single node, or through the network in a cluster.  To cut down the overall overhead, first we try to reduce the amount of data to be transferred. Further more, we try to parallelize the computation and communication, in order to mask the communication time. 
Here we discuss synchronous training only (i.e., a single worker group),
which has the identical theoretical convergence as training in a single worker.
Optimization techniques that may affect convergence rate of SGD are not considered, e.g., asynchronous SGD (i.e., multiple worker groups) and parameter compression~\cite{DBLP:conf/interspeech/SeideFDLY14}.
The following analysis works for training either over multiple CPU nodes or over multiple GPU cards on a single node.

\subsubsection{Reducing Data Transferring}\label{sec:reduce-comm}
Corresponding to the three basic partitioning strategies, there are three sources of communication overhead. The first partitioning strategy results in data being transferred along the boundary layers between workers. To reduce the overhead, we can select the boundary layers at ``low traffic'' positions. In other words, layers with smaller feature dimensions are preferred, as they pass less data to the destination layer. The second partitioning strategy, i.e., data parallelism, replicates the parameters for each layer, hence their gradients are transferred to a central parameter server for aggregation, and the new parameter values are broadcast back for the next iteration. To reduce the overhead, we can apply data parallelism on layers with fewer parameters. The third partitioning strategy, i.e., model parallelism, slices the feature vector into sub-vectors, hence some layers (e.g., the fully connected layer whose neuron depends on all neurons of its source layer) need to fetch sub-vectors of source layers from other workers to compute its own feature vector. To address this, we can apply model parallelism only for those layers whose 
neuron dependency is element-wise or with small feature dimension.
For the last partitioning strategy, i.e.,  hybrid partitioning, we can compare the overall overhead of different combinations of the basic partitioning strategies and select the combination that incurs minimal overhead. To illustrate, we use the popular benchmark model, i.e., AlexNet, as an example. AlexNet is a feed-forward model with single path, the $i^{th}$ layer depends on $(i-1)^{th}$ layer directly. It is not feasible to parallelize subsets of layers as in MDNN, therefore we do not consider the first partitioning strategy. Next, we discuss every type of layer involved in AlexNet one by one. 

Convolution layers contain 5\% of the total parameters but 90-95\% of the computation, according to AlexNet~\cite{DBLP:journals/corr/Krizhevsky14}.
It is essential to distribute the computation from these layers.
Considering that convolution layers have large feature dimensions, it is natural to apply data parallelism.

\begin{figure}[ht]
	\centering
	\subfigure[Two fully connected layers.\label{fig:fc}]{ %
		\includegraphics[width=.32\textwidth]{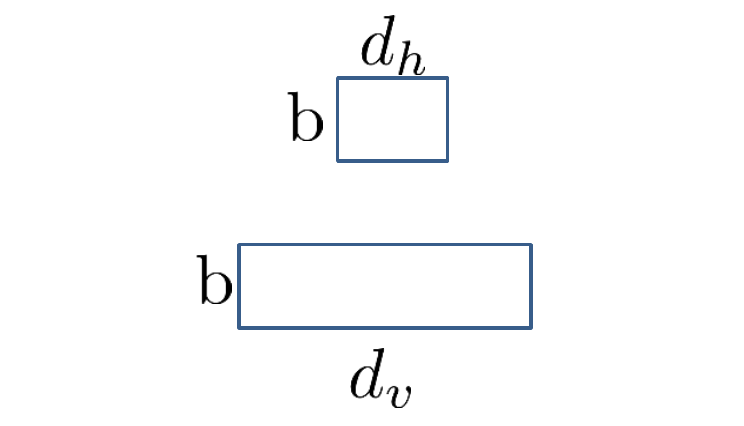}}
	\subfigure[Partition on hidden layer.\label{fig:fc-hid}]{ %
		\includegraphics[width=.32\textwidth]{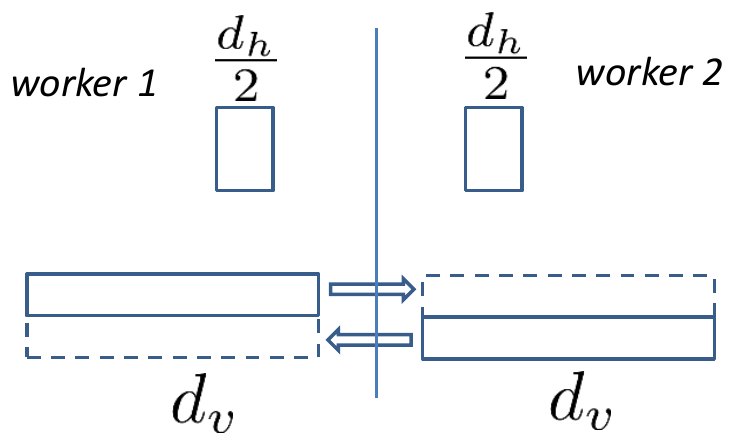}}
	\subfigure[Partition on visible layer.\label{fig:fc-vis}]{ %
		\includegraphics[width=.32\textwidth]{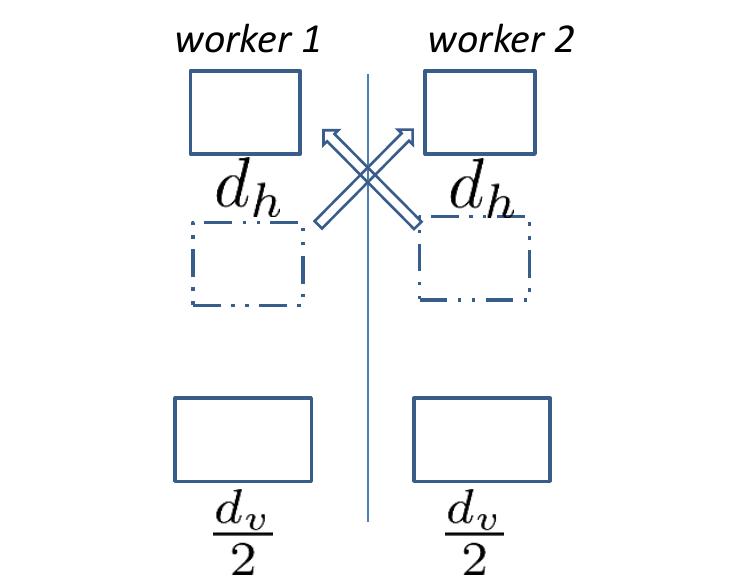}}
	\caption{Distributed computing for fully connected layers.}
\end{figure}

Fully connected layers occupy 95\% of the total parameters and 5-10\% of computation~\cite{DBLP:journals/corr/Krizhevsky14}, therefore we  avoid data parallelism for them.
Particularly, with data parallelism, the communication overhead per worker is $p$, where $p$ is the size of the (replicated) parameters. Let $b$ be the effective mini-batch size (summed over all workers), $K$ be the number of workers, and $d_v$ (resp. $d_h$) be the length of the visible (resp. hidden) feature vector. Using model parallelism, the communication overhead per worker is $b*d_v$  for Figure~\ref{fig:fc-hid}, including sending its own data to other workers, i.e., $b*d_v/K $, and receiving data from other workers, i.e., $b*(K-1)*d_v/K$.
For the case in Figure~\ref{fig:fc-vis}, the overhead is $b*d_h$, where each worker computes the partial feature of the hidden layer using sub-vectors of the visible layer and concatenates them to get the complete feature.
To compare the two strategies, data parallelism is costlier than model parallelism when $p>b*d_v$ or $p>b*d_h$.
For the first fully connected layer in AlexNet, $p$ is about 177 million while $d_v=d_h=4096$.
In practice, there are fewer than $K=8$ GPU cards on a single node (i.e., $K<=8$), and each worker runs with fewer than $128$ samples per mini-batch, thus data parallelism is much costlier than model parallelism. Another approach is no-partitioning for these fully connected layers. The overhead, in this case, comes from transferring features with other workers, which is $b*(K-1)*d_v/K$. However, the computation power is reduced to $1/K$ as the workload of $K$ workers is conducted by a single worker. 

For pooling layers and local responsive normalization layers, each neuron depends on many neurons from their source layers. Moreover, they are inter-leaved with convolution layers, thus it is cheaper to apply data parallelism than model parallelism for them. For the remaining layers, they do not have parameters and their neurons depend on source neurons element-wise, hence their partitioning strategies just need to be consistent with their source layers. 
Consequently, a simple hybrid partitioning strategy for AlexNet~\cite{DBLP:journals/corr/Krizhevsky14} can be to apply data parallelism for layers before (or under) the first fully connected layer, and then apply model parallelism or no parallelism for all other layers. The above hybrid partitioning can be easily configured in SINGA by setting the \emph{partition\_dim} of each layer correspondingly. 

\subsubsection{Overlapping Computation and Communication}\label{sec:comp-comm}
\begin{figure}
	\centering
	\includegraphics[width=.6\textwidth]{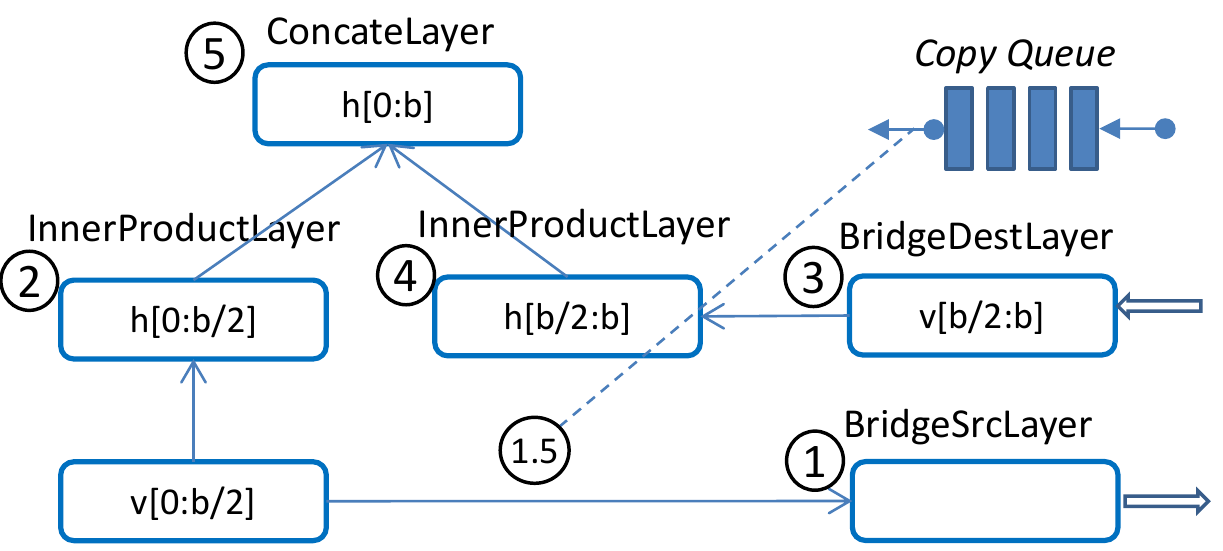}
	\caption{Parallelize computation and communication for a GPU worker.}
	\label{fig:comp-comm}
\end{figure}

Overlapping the computation and communication is another common technique for system optimization. In SINGA, the communication comprises transferring parameter gradients and values, and transferring layer data and gradients. 
First, for parameter gradients/values, we can send them asynchronously while computing other layers.
Take Figure~\ref{fig:mlp} as an example, after the hidden layer finishes \emph{ComputeFeature}, we can send the gradients asynchronously to the server for updates while the worker continues to load data for the next iteration.
Second, the transferring of layer data/gradients typically comes from model partitioning as discussed in Section~\ref{sec:reduce-comm}.
In this case, each worker owns a small subset of data and fetches all rest from other workers.
To overlap the computation and communication, each worker can just initiate the communication and then compute over its own data asynchronously.
Take the Figure~\ref{fig:fc-hid} as an example, to parallelize the computation and communication, SINGA runs over the layers shown in Figure~\ref{fig:comp-comm} in order. The \emph{BridgeSrcLayer::ComptueFeature} initiates the sending operations and returns immediately. The \emph{BridgeDestLyer::ComputeFeature} waits until data arrives (by checking a signal for the ending of data transferring). All layers are sorted in topology order followed by communication priority.

For training with multiple GPUs on a single node  we need another mechanism to overlap data transfer and computation.
The data transfer from CPU to GPU can only be processed by the worker itself.
Consequently, as in Figure~\ref{fig:comp-comm}, the BridgeDestLayer has to copy the data from CPU to GPU itself instead of relying on other threads to do it.
In order to execute the copy operation synchronously, we add a data copy queue for each GPU worker.
This queue is checked frequently (e.g., before visiting each layer) to initiate the copy operation (from CPU to GPU) asynchronously.
A copy event is pushed into the queue by other threads (e.g., stub or other worker).
A callback function is associated with each copy event to signal the end of the copy operation, i.e., data transferring.
If the copy event for the BridgeDestLayer is initiated after step 1 (say at step 1.5), it could be done in parallel with step 2. 
Later, when the worker visits BrdigeDestLayer (i.e., step 3), the copy event could have already finished. The transferring of parameter values/gradients is processed in the same way. Each worker initiates asynchronous sending operations to servers immediately after it gets the gradients.
After updating, the servers enqueue the event for copying the fresh parameter values back to the workers. Then the workers can parallelize the copy operations and computation for other layers.
Depending on the TrainOneBatch algorithm, we may assign a different priority for each copy event. For example, for the BP algorithm, the fresh parameters of the bottom layers may have higher priority because the bottom layers will be visited earlier than other layers in the next iteration. Otherwise, the computation of the bottom layers would be blocking while it waits for the fresh parameter. 
\section{Experimental Study}\label{sec:experiment}

We evaluated SINGA with real-life multimedia applications. Specifically, we used SINGA to train the models discussed in
Section~\ref{sec:app}, which required little development effort since SINGA comes with many built-in layers and
algorithms. We then measured SINGA's training performance in terms of efficiency and scalability when running on CPUs
and GPUs. We found that SINGA is more efficient than other open-source systems, and it is scalable for both synchronous
and asynchronous training.

\subsection{Applications of SINGA}
We trained models for the example applications in Section~\ref{sec:app} using SINGA. Users can train these models following the instructions on-line\footnote{\url{http://singa.apache.org/docs/examples.html}}.The neural nets are configured using the built-in layers as shown in Figure~\ref{fig:mdnn-net}, \ref{fig:rbm-net}, \ref{fig:rnn-net}.

\begin{figure}
	\centering
	\includegraphics[width=.75\textwidth]{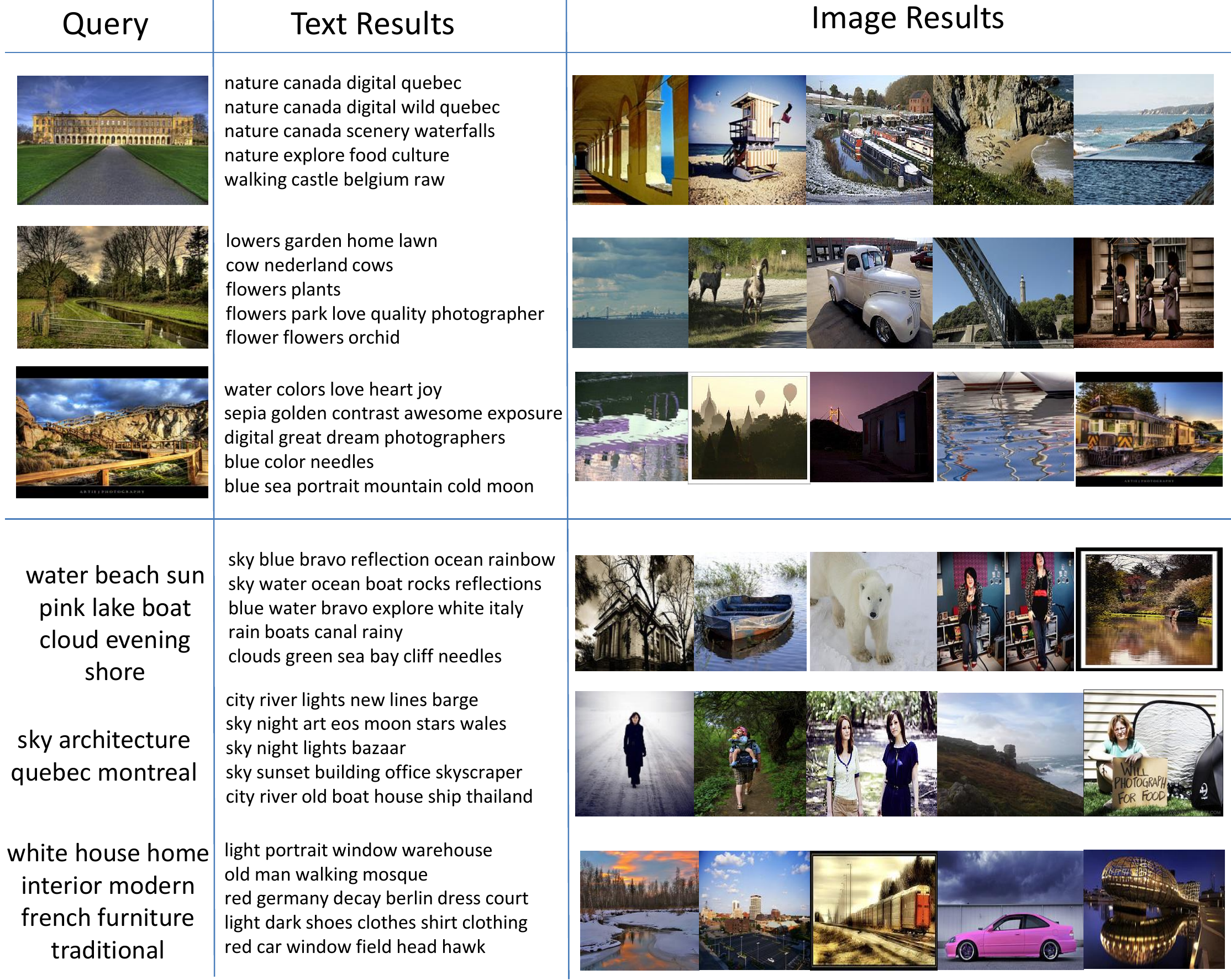}
	\caption{Multi-Modal Retrieval. Top 5 similar text documents (one line per document) and images are displayed.}
	\label{fig:mdnn-sample}
\end{figure}

\textbf{Multi-modal Retrieval}.
We trained the MDNN model for multi-modal retrieval application. We used NUS-WIDE dataset~\cite{nus-wide-civr09}, which has roughly 180,000 images after removing images without tags or from non-popular categories.  Each image is associated with several tags. We used Word2Vec~\cite{DBLP:conf/nips/MikolovSCCD13} to learn a word embedding for each tag and aggregated the embedding of all the tags from the same image as a text feature. Figure~\ref{fig:mdnn-sample} shows sample search results. We first used images as queries to retrieve similar images and text documents. It can be seen that image results are more relevant to the queries. For instance,  the first image result of the first query is relevant because both images are about architecture, but the text results are not very relevant. This can be attributed to the large semantic gap between different modalities, making it difficult to locate semantically relevant objects in the latent (representation) space.

\begin{figure}
	\centering
	\subfigure[Bottom RBM weight matrix.\label{fig:rbm-weight}]{ %
		\includegraphics[width=.4\textwidth]{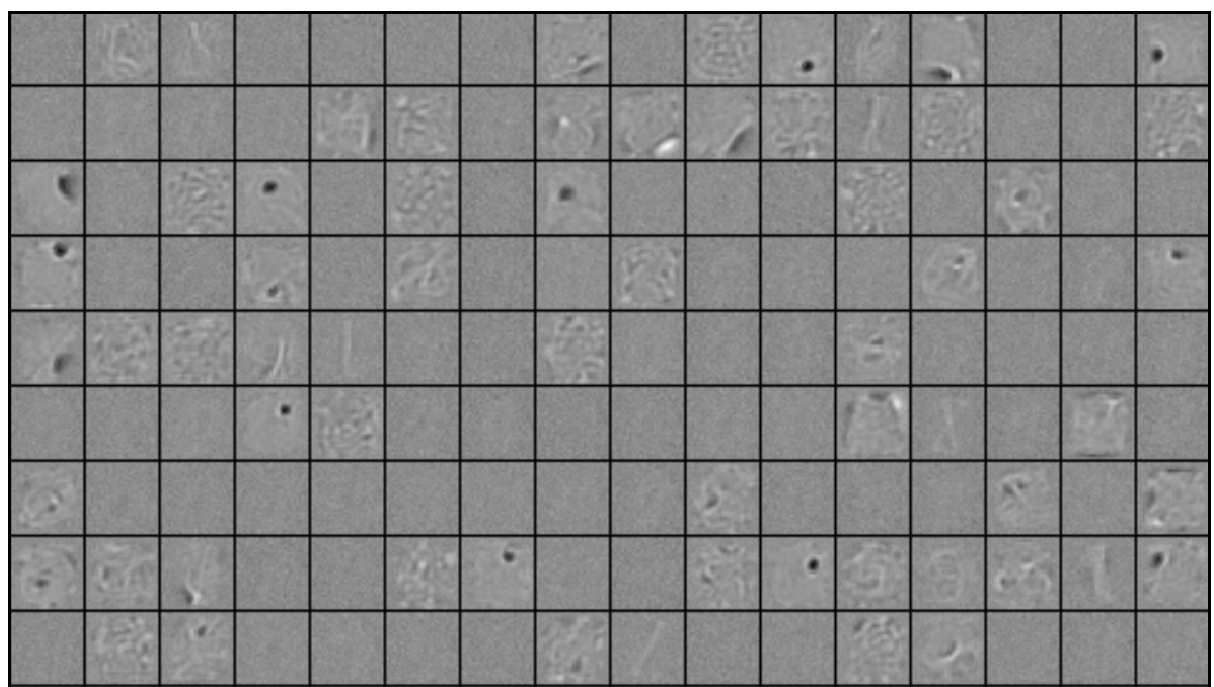}}
	\hfill
	\subfigure[Top layer features.\label{fig:auto-encoder-fea}]{ %
	  	\includegraphics[width=.37\textwidth]{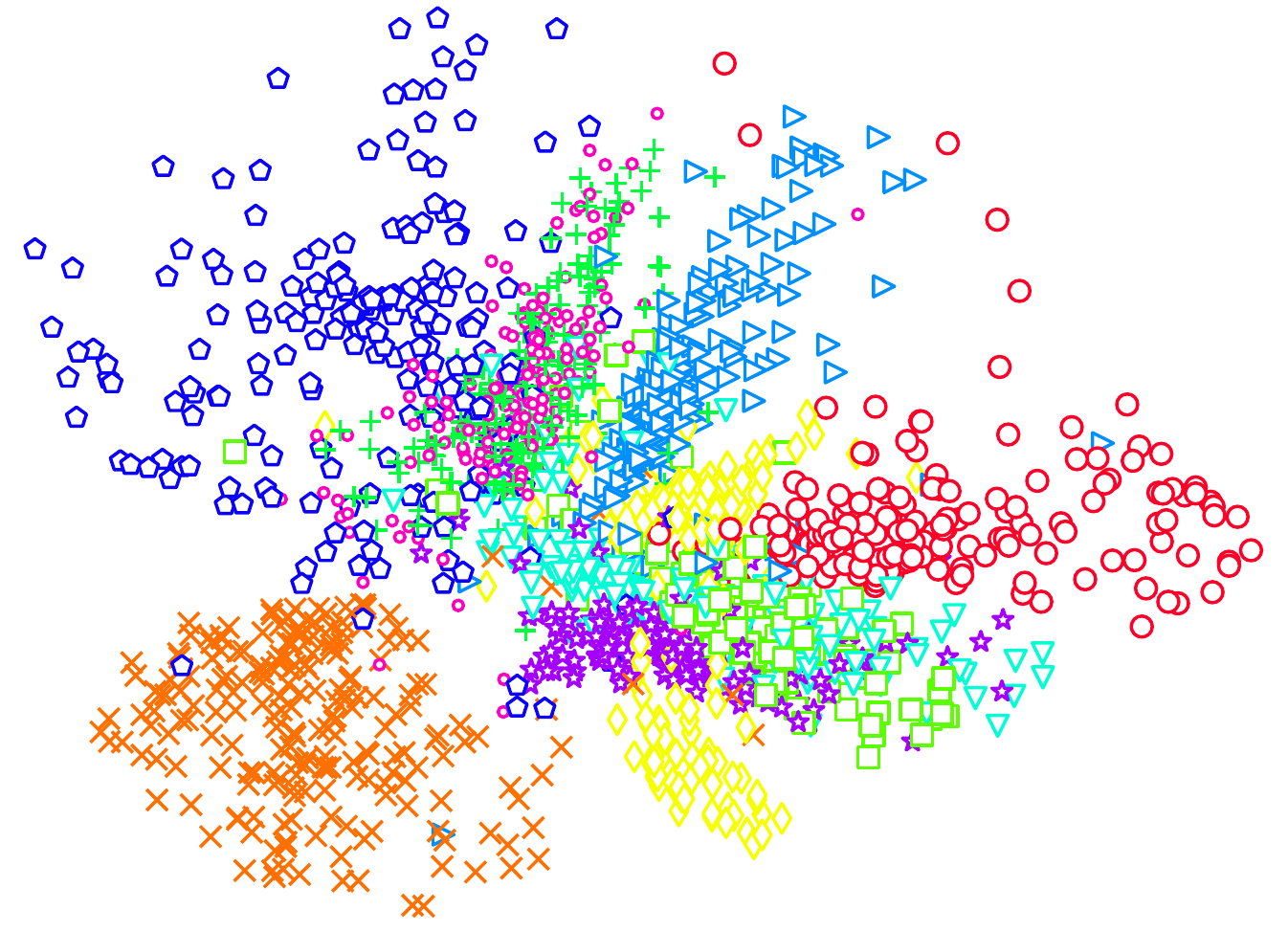} }
	\caption{Visualization of the weight matrix in the bottom RBM and top layer features in the deep auto-encoder.}
\end{figure}

\textbf{Dimensionality Reduction}.
We trained RBM models to initialize the deep auto-encoder for dimensionality reduction. We used the MNIST\footnote{http://yann.lecun.com/exdb/mnist/} dataset consisting of 70,000 images of hand-written digits. Following the configuration used in \cite{HinSal06}, we set the size of each layer as 784$\rightarrow$1000$\rightarrow$500$\rightarrow$250$\rightarrow$2. Figure~\ref{fig:rbm-weight} visualizes sample columns of the weight matrix of the bottom (first) RBM. We can see that Gabor-like filters are learned. Figure~\ref{fig:auto-encoder-fea} depicts the features extracted from the top-layer of the auto-encoder, wherein one point represents one image. Different colors represent different digits. We can see that most images are well clustered according to the ground truth, except for images of digit '4' and '9' (central part) which have some overlap (in practice, handwritten '4' and '9' digits are fairly similar in shape).

\begin{figure}[ht]
	\centering
	\subfigure[Training accuracy\label{fig:char-rnn-acc}.]{ %
		\includegraphics[width=.4\textwidth]{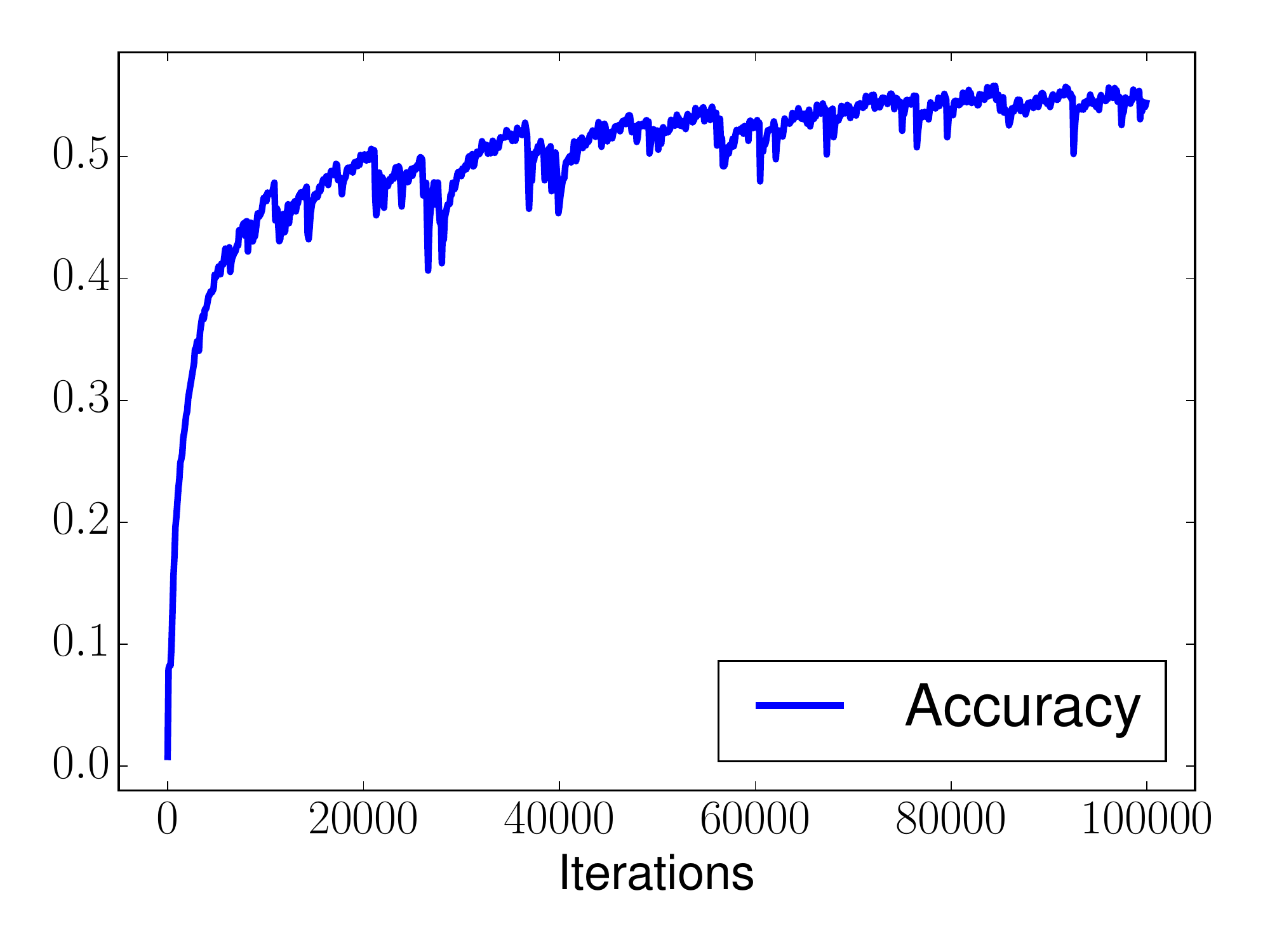}}
	\subfigure[\label{fig:char-rnn-loss} Training loss.]{ %
		\includegraphics[width=.4\textwidth]{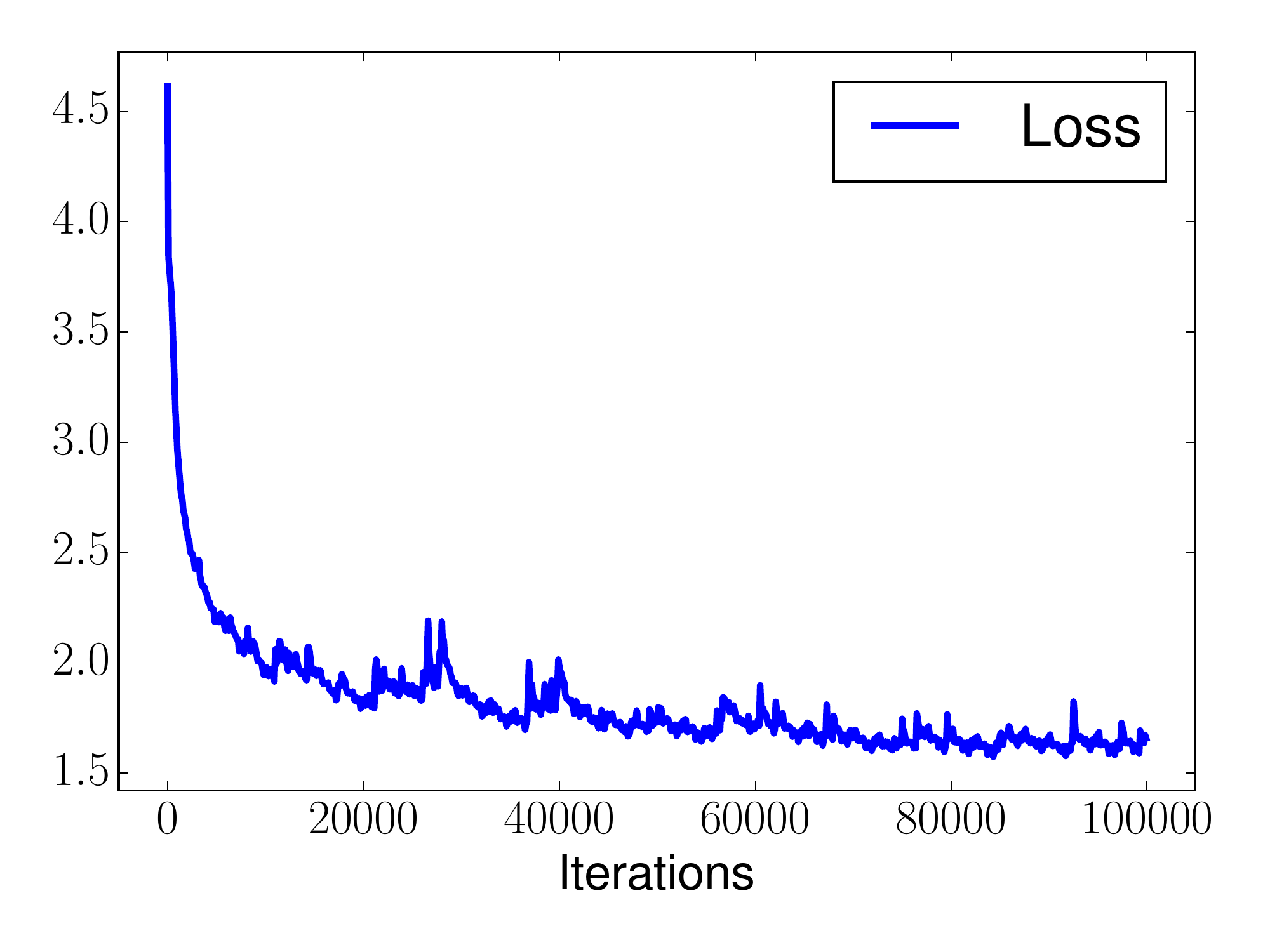}}
	\caption{Training accuracy and loss of Char-RNN.}
\end{figure}

\textbf{Char-RNN}
We used the Linux kernel source code extracted using an online
script\footnote{\url{http://cs.stanford.edu/people/karpathy/char-rnn}} for this application. The dataset is about 6 MB.
The RNN model is configured similar to Figure~\ref{fig:rnn-net}. Since this dataset is small, we used one stack of
recurrent layers (Figure~\ref{fig:rnn-net} has two stacks). The training loss and accuracy is shown in
Figure~\ref{fig:char-rnn-loss}. We can see that the Char-RNN model can be trained to predict the next character given
previous characters in the source code more and more accurately.

\subsection{Training Performance Evaluation on CPU}
We evaluated SINGA's training efficiency and scalability for both synchronous and asynchronous frameworks on a single
multi-core node, and on a cluster of commodity servers.

\subsubsection{Methodologies}
The deep convolution neural network\footnote{https://code.google.com/p/cuda-convnet/} for image classification was used as the training model for benchmarking.
The training was conducted over the CIFAR10 dataset\footnote{http://www.cs.toronto.edu/~kriz/cifar.html} which has
50,000 training images and 10,000 test images. 

For the single-node setting, we used a 24-core server with 500GB memory. The 24 cores are distributed into 4 NUMA
nodes (Intel Xeon 7540). Hyper-threading is turned on. For the multi-node setting, we used a 32-node cluster. Each
cluster node is equipped with a quad-core Intel Xeon 3.1 GHz CPU and 8GB memory. The cluster nodes are connected by
a 1Gbps switch.

SINGA uses Mshadow\footnote{https://github.com/dmlc/mshadow} and OpenBlas\footnote{http://www.openblas.net/} to
accelerate linear algebra operations (e.g., matrix multiplication). Caffe's im2col and pooling
code~\cite{jia2014caffe} is adopted to accelerate the convolution and pooling operations. We compiled SINGA using GCC with
optimization level O2.

\subsubsection{Synchronous training}
We compared SINGA with CXXNET\footnote{https://github.com/dmlc/cxxnet} and Caffe~\cite{jia2014caffe}.  All three systems
use OpenBlas to accelerate matrix multiplications. Both CXXNET and Caffe were compiled with their default optimization
levels: O3 for the former and O2 for the latter.  We observed that because synchronous training has the same convergence
rate as that of sequential SGD, all systems would converge after same number of iterations (i.e., mini-batches). This
means the difference in total training time among these systems is attributed to the efficiency of a single iteration.
Therefore, we only compared the training time for one iteration. We ran 100 iterations for each system and averaged the
result time over 50 iterations: $30^{\text{th}}$ to $80^{\text{th}}$ iteration, in order to to avoid the effect of
starting and ending phases.

\begin{figure}
	\centering
	\subfigure[On the single node\label{fig:sync}.]{ %
		\includegraphics[width=.4\textwidth]{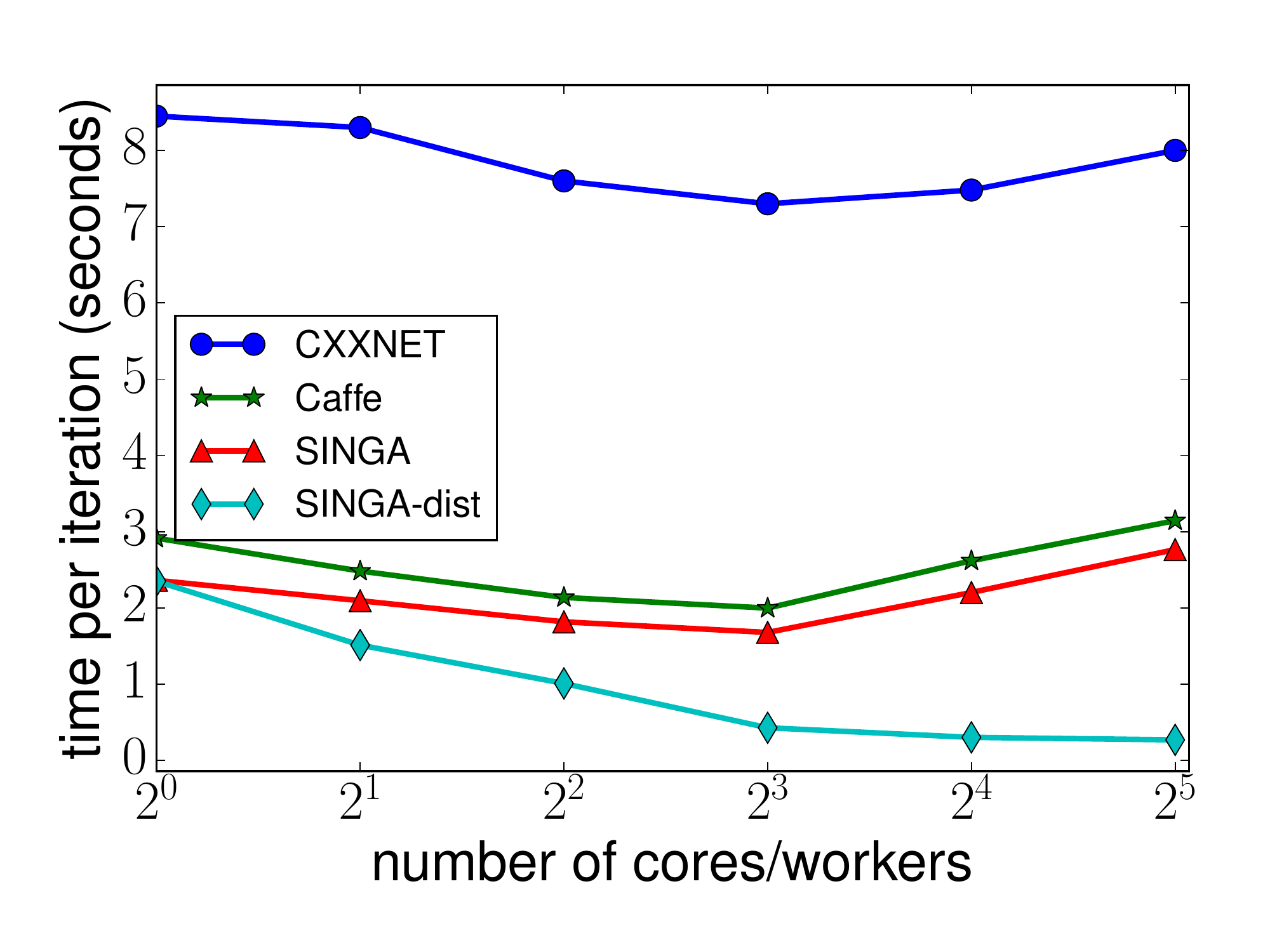}}
	\subfigure[\label{fig:dist-sync} On the 32-node cluster.]{ %
		\includegraphics[width=.4\textwidth]{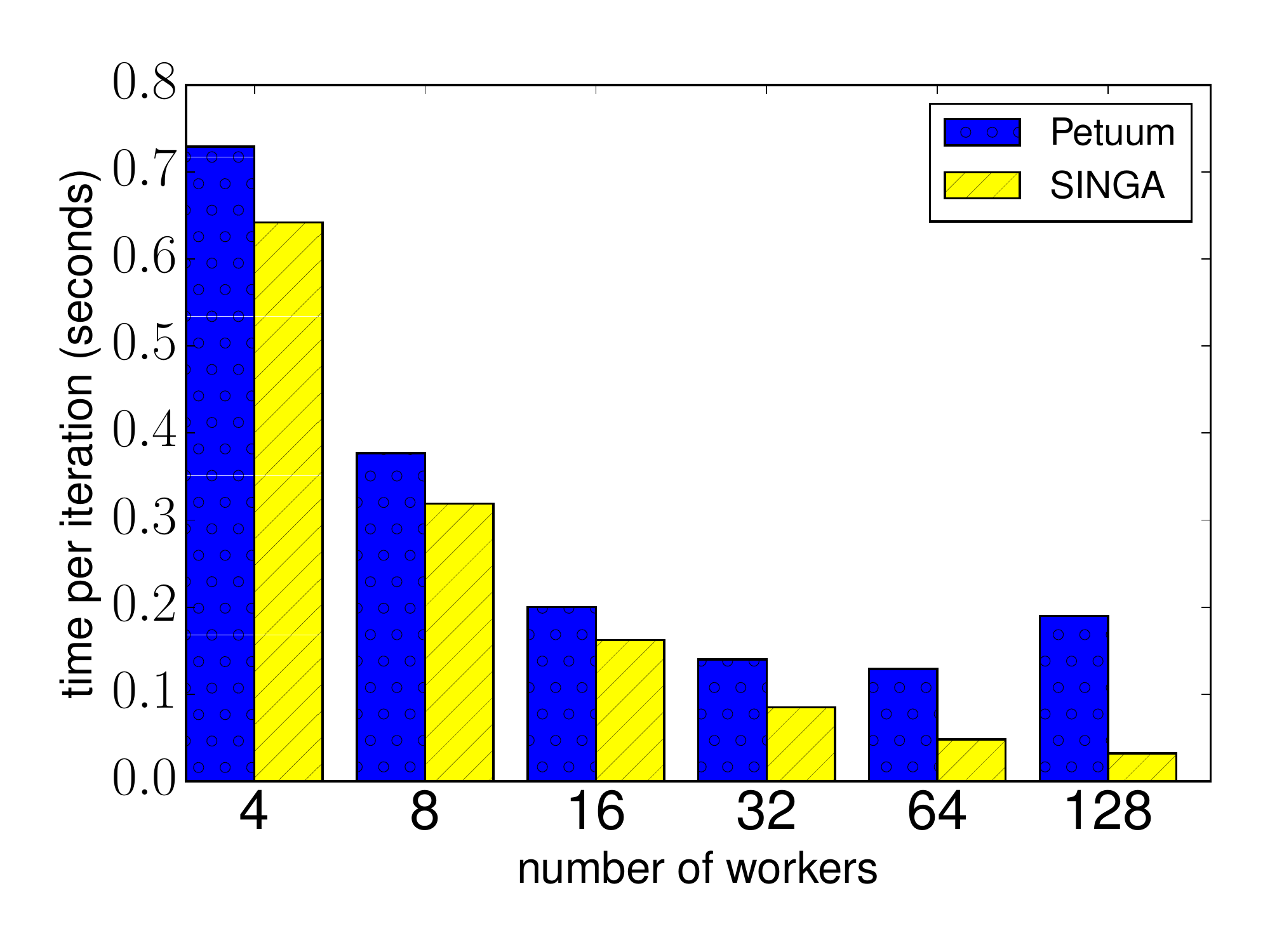}}
	\caption{Synchronous training.}
\end{figure}

On the 24-core single node, we used 256 images per mini-batch and varied the number of OpenBlas's threads. The result is
shown in Figure~\ref{fig:sync}. {\em SINGA-dist} represents the SINGA configuration in which there are multiple workers,
each worker has 1 OpenBlas thread\footnote{OPENBLAS\_NUM\_THREADS=1}. In contrast,  {\em SINGA} represents the
configuration which has only 1 worker. We configured SINGA-dist with the cluster topology consisting of  one server
group with four servers and one worker group with varying number of worker threads (Figure~\ref{fig:sync}).
In other words, SINGA-dist ran as the in-memory Sandblaster framework.  We can see that SINGA-dist has the best
overall performance: it is the fastest for each number of threads, and it is also the most scalable.  Other systems using
multi-threaded OpenBlas scale poorly. This is because OpenBlas has little awareness of the application, and
hence it cannot be fully optimized.  For example, it may only parallelize specific operations such as large matrix
multiplications. In contrast, in SINGA-dist partitions the mini-batch equally between workers and achieves parallelism
at the worker level.  Another limitation of OpenBlas, as shown in Figure~\ref{fig:sync}, is that when there were more
than $8$ threads, the overheads caused by cross-CPU memory access started to have negative effect on the overall
performance.

\begin{figure}
	\centering
	\subfigure[\label{fig:caffe-async} Caffe on the single node.]{ %
		\includegraphics[width=.32\textwidth]{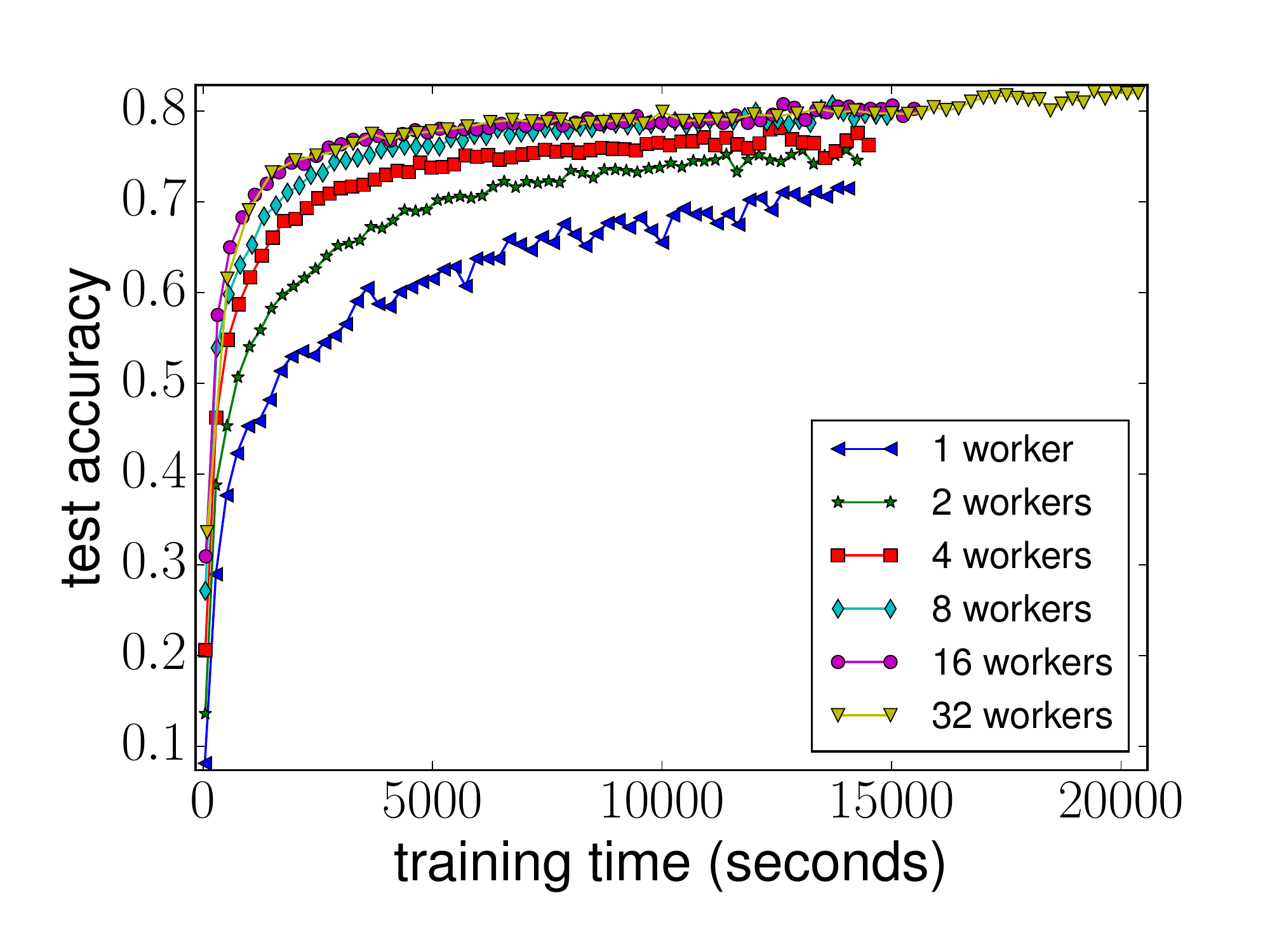}}
	\subfigure[\label{fig:singa-async} SINGA on the single node.]{ %
		\includegraphics[width=.32\textwidth]{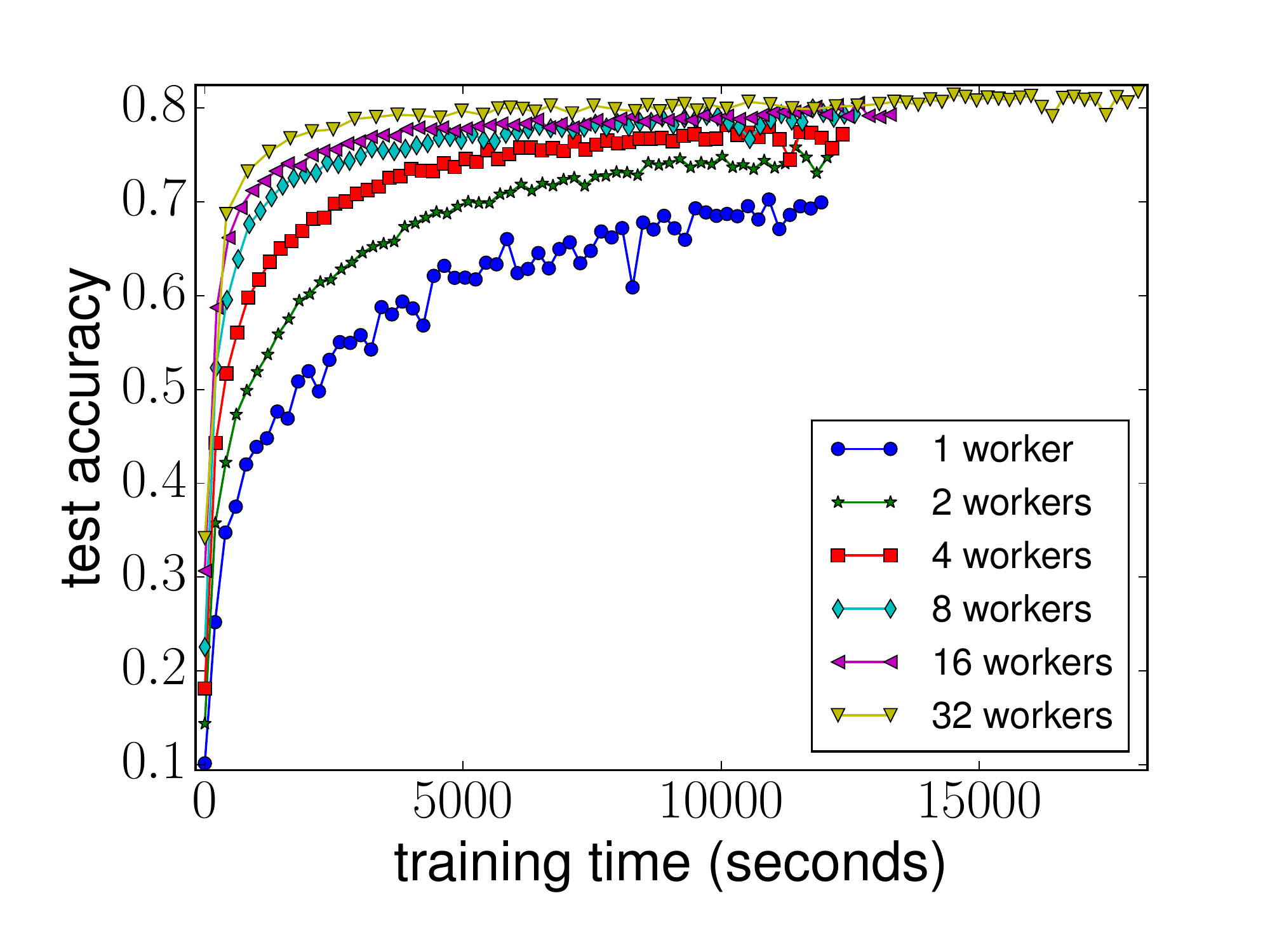}}
	\subfigure[\label{fig:singa-async-dist}SINGA on the 32-node cluster.]{ %
		\includegraphics[width=.32\textwidth]{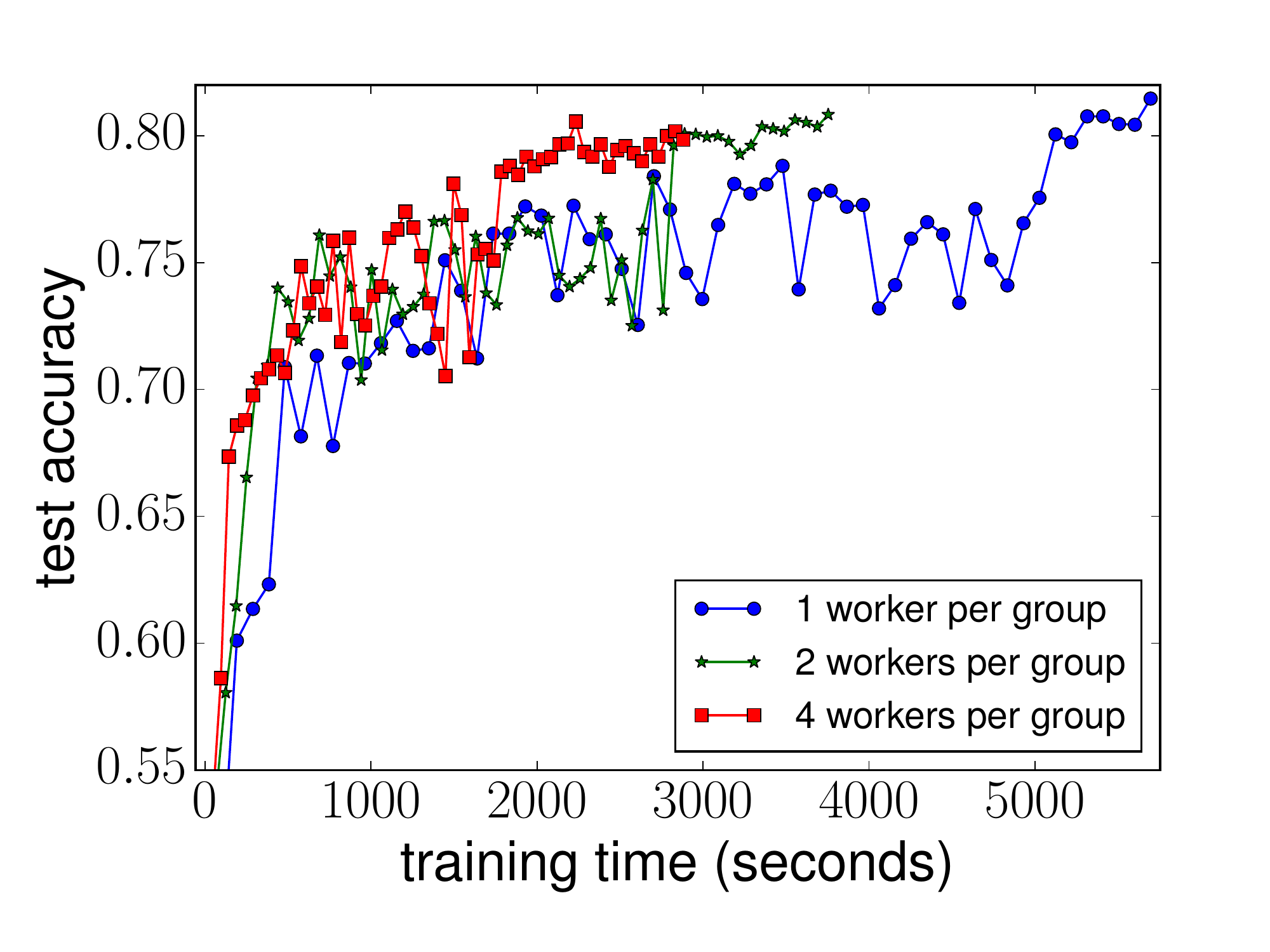}}
	\caption{Asynchronous training.}
\end{figure}

On the 32-node cluster, we compared SINGA against another distributed machine learning framework called
Petuum~\cite{DBLP:journals/corr/DaiWZKLYHX13}. Petuum runs Caffe as an application to train deep learning models. It
implements a parameter server to perform updates from workers (clients), while the workers run synchronously. We used a
larger mini-batch size (512 images) and disabled OpenBlas multi-threading. We configured SINGA's cluster topology to realize
the AllReduce framework: there is 1 worker group and 1 server group, and in each node there
are 4 workers and 1 server.  We varied the size of worker group from 4 to 128, and the server group size from 1 to 32.
We note that one drawback of synchronous distributed training is that it cannot scale to too many nodes because
there is typically an upper limit on the mini-batch size (1024 images, for instance).  Consequently, there is an upper bound on the
number of workers we can launch (1024 workers, for instance), otherwise some workers will not be assigned any image to train.
Figure~\ref{fig:dist-sync} shows that SINGA achieves almost linear scalability. In contrast, Petuum scales up to 64
workers, but becomes slower when 128 workers are launched. It might be attributed to the communication overheads at the
parameter server and the synchronization delays among workers.

\subsubsection{Asynchronous training}
We compared SINGA against Caffe which has support for in-memory asynchronous training. On the single node, we configured
Caffe to use the in-memory Hogwild~\cite{DBLP:conf/nips/RechtRWN11} framework, and SINGA to use the in-memory Downpour
framework.  Their main difference is that parameter updates are done by workers in Caffe and by a single server (thread)
in SINGA.  Figure~\ref{fig:caffe-async} and Figure~\ref{fig:singa-async} show the model accuracy versus training time
with varying numbers of worker groups (i.e. model replicas). Every worker processed 16 images per iteration, for a total
of 60,000 iterations.  We can see that SINGA trains faster than Caffe. Both systems scale well as the number of
workers increases, both in terms of the time to reach the same accuracy and of the final converged accuracy. We can also
observe that the training takes longer with more workers.  This is due to the increased overhead
 in context-switching when there are more threads (workers). Finally, we note from the results that the performance
 difference becomes smaller when the cluster size (i.e., the number of model replicas) reaches 16. This implies that
 there would be little benefit in having too many model replicas. Thus, we fixed the number of model replicas (i.e.,
 worker groups) to $32$ in the following experiments for the distributed asynchronous training.

On the 32-node cluster, we used mini-batch of 16 images per worker group and 60,000 training iterations. We varied the
number of workers within one group, and configured the distributed
Downpour framework to have 32 worker groups and 32 servers per server group (one server thread per
node). We can see from Figure~\ref{fig:singa-async-dist} that with more workers, the training is faster because each worker processes fewer images. However, the training is not as stable as in the single-node setting. This may be
caused by the delay of parameter synchronization between workers, which is not present in single-node training because
parameter updates are immediately visible on the shared memory.  The final stage of training (i.e., last few
points of each line) is stable because there is only one worker group running during that time, namely the testing
group. We note that using a warm-up stage, which trains the model using a single worker group at the beginning, may help
to stabilize the training as reported in Google's DistBelief system~\cite{DBLP:conf/nips/DeanCMCDLMRSTYN12}.

\subsection{Training Performance Evaluation on GPU}
We evaluated the training performance of SINGA running on GPUs. We first analyzed the two optimization techniques
discussed in  Section~\ref{sec:opt}, then we compared  SINGA with other open source, state-of-the-art systems. 

\subsubsection{Methodologies}
We used the online benchmark model\footnote{\url{https://github.com/soumith/convnet-benchmarks}} as the training
workload. The model is adapted from the AlexNet~\cite{DBLP:journals/corr/Krizhevsky14} model with some layers omitted.
We performed experiments on single node which has one Intel i7 k5820 CPU, 3 GTX 970 GPUs and 16 GB memory.  CUDA v7.0 and cuDNN v3 are used. We measured
the performance of synchronous training with different optimization techniques over 1, 2 and 3 GPU cards, and compared
that with other state-of-the-art systems with the same configuration.

\begin{figure}
	\centering
	\subfigure[Overlapping communication and computation. \label{fig:comp-comm-exp}]{ %
		\includegraphics[width=.45\textwidth]{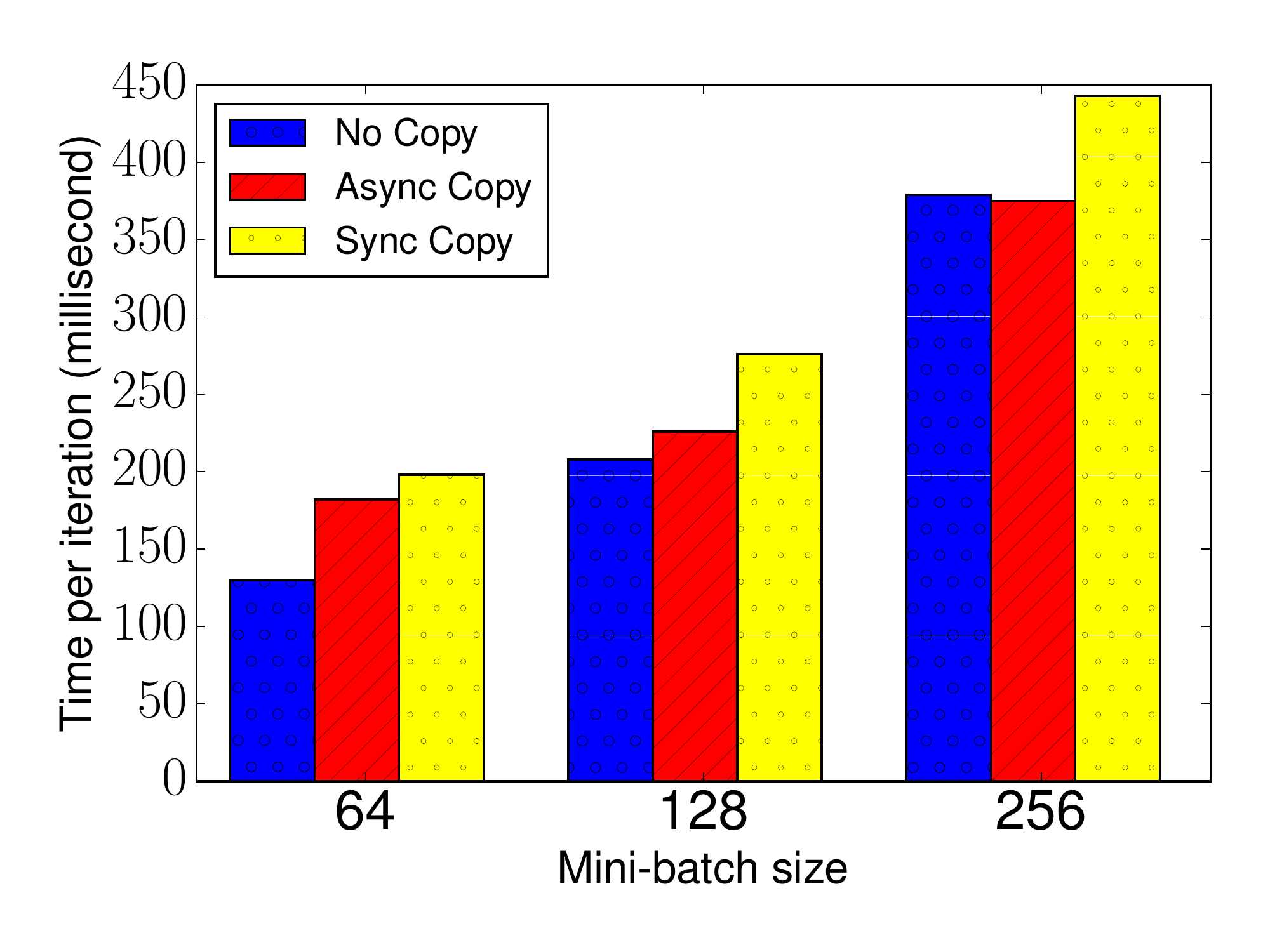}}
	\subfigure[Reducing transferred data.\label{fig:reduce-data-exp}]{ %
		\includegraphics[width=.44\textwidth]{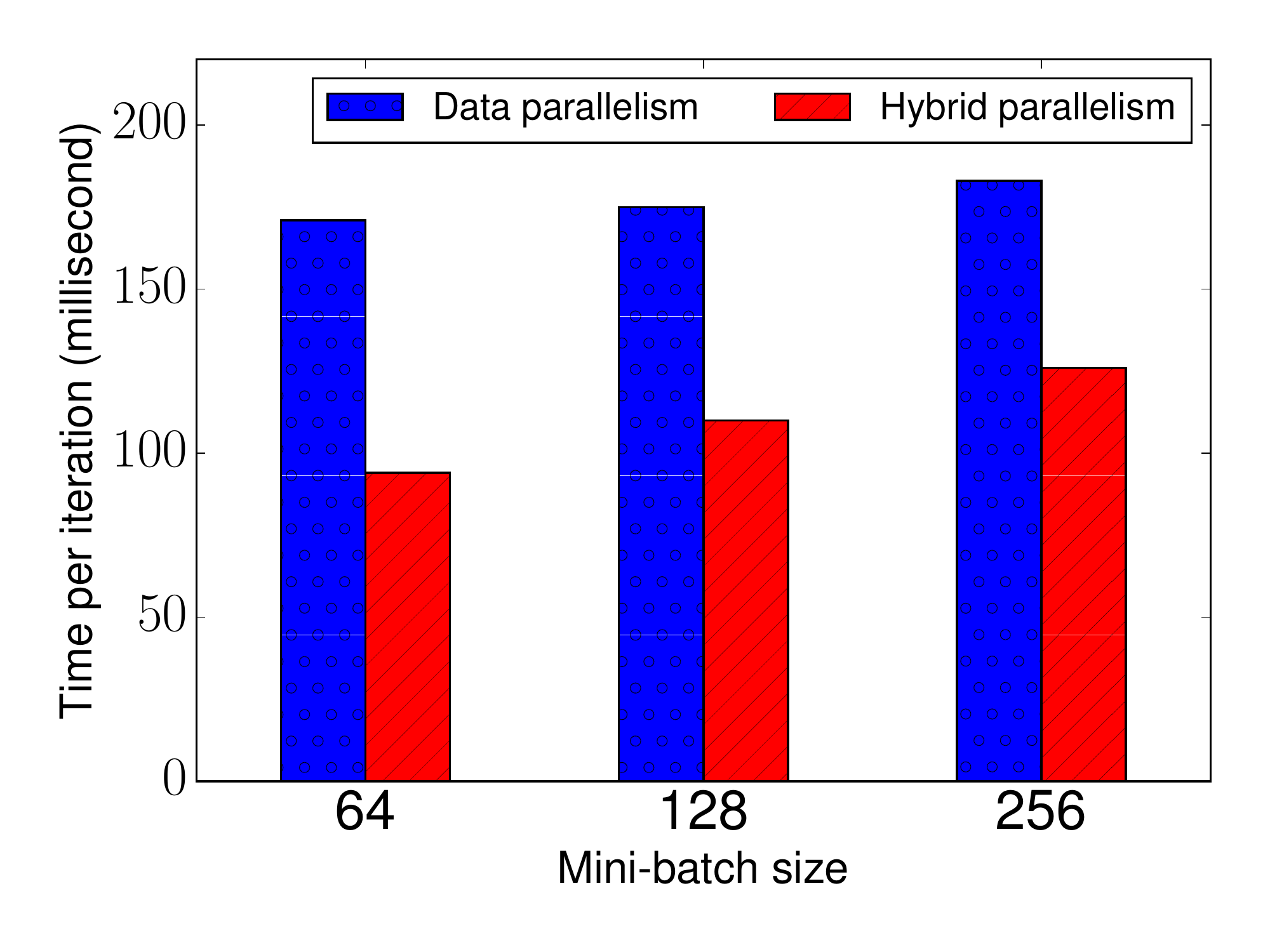}}
	\caption{Effect of optimization techniques.}
\end{figure}

\subsubsection{Overlapping Communication and Computation}

In Section~\ref{sec:comp-comm}, we analyzed the optimization technique for hiding the communication overhead by overlapping it with the computation. Here we evaluate the effect of this technique. Particularly, we compare the efficiency in terms of time per iteration for three versions of SINGA. No Copy version indicates that there is no communication between GPU and CPU, which is widely used for training with a single GPU, where all SGD operations including parameter update are conducted on the single GPU. The other two versions conduct BP algorithm on GPU and parameter updating on CPU, differing only by whether the GPU and CPU communicate synchronously or asynchronously.

Figure~\ref{fig:comp-comm-exp} shows the time per iteration with different mini-batch size. First, we can see that No Copy is the fastest one because it has no communication cost at all. Second, Async Copy is faster than Sync Copy, which suggests that the asynchronous data transferring benefits from the overlapping communication and
computation. Moreover, we can see that when the mini-batch increases, the difference between Async Copy and Sync Copy decreases. This is because for large mini-batches, the BP algorithm spends more time doing computation, which increases 
the overlap area of computation and communication, effectively reducing the overhead. For mini-batch size = 256, Async Copy is even faster than No Copy, this is because Async Copy does not do parameter update, which is done by the server in parallel with BP. However, No Copy has to do BP and parameter updating in sequential.

\subsubsection{Reducing Data Transferring}

In Section~\ref{sec:reduce-comm}, we discussed how hybrid partitioning is better than other  strategies in terms of the
overheads in transferring feature vectors between layers in different workers. To demonstrate its effectiveness, we ran
SINGA using two partitioning strategies, i.e., data partitioning and hybrid partitioning for the first fully connected layer in AlexNet.
Figure~\ref{fig:reduce-data-exp} shows the time per iteration with different mini-batch sizes. We can see that  
hybrid partitioning has better performance over data partitioning and single GPU training. For data partitioning, only parameter gradients and values are transferred, which is independent of the mini-batch size, thus the time per iteration does not change much when mini-batchs size increases. For hybrid partitioning, when the mini-batch size increases, more feature vectors would be transferred. Hence, the time increases.

\subsubsection{Comparison with Other Systems}
We compared SINGA with four other state-of-the-art deep learning systems, namely Torch~\cite{collobert:2011c},
Caffe~\cite{jia2014caffe}, TensorFlow~\cite{tensorflow2015-whitepaper} and MxNet~\cite{chen2015mxnet}. 
For fair comparison, we turned off the manual tuning option provided by cuDNN and we use AllreduceCPU for MxNet which aggregates the gradients on CPU like SINGA.

\begin{figure}
	\centering
	\subfigure[\label{fig:fix-per-worker} Mini-batch size per worker = 96.]{ %
		\includegraphics[width=.37\textwidth]{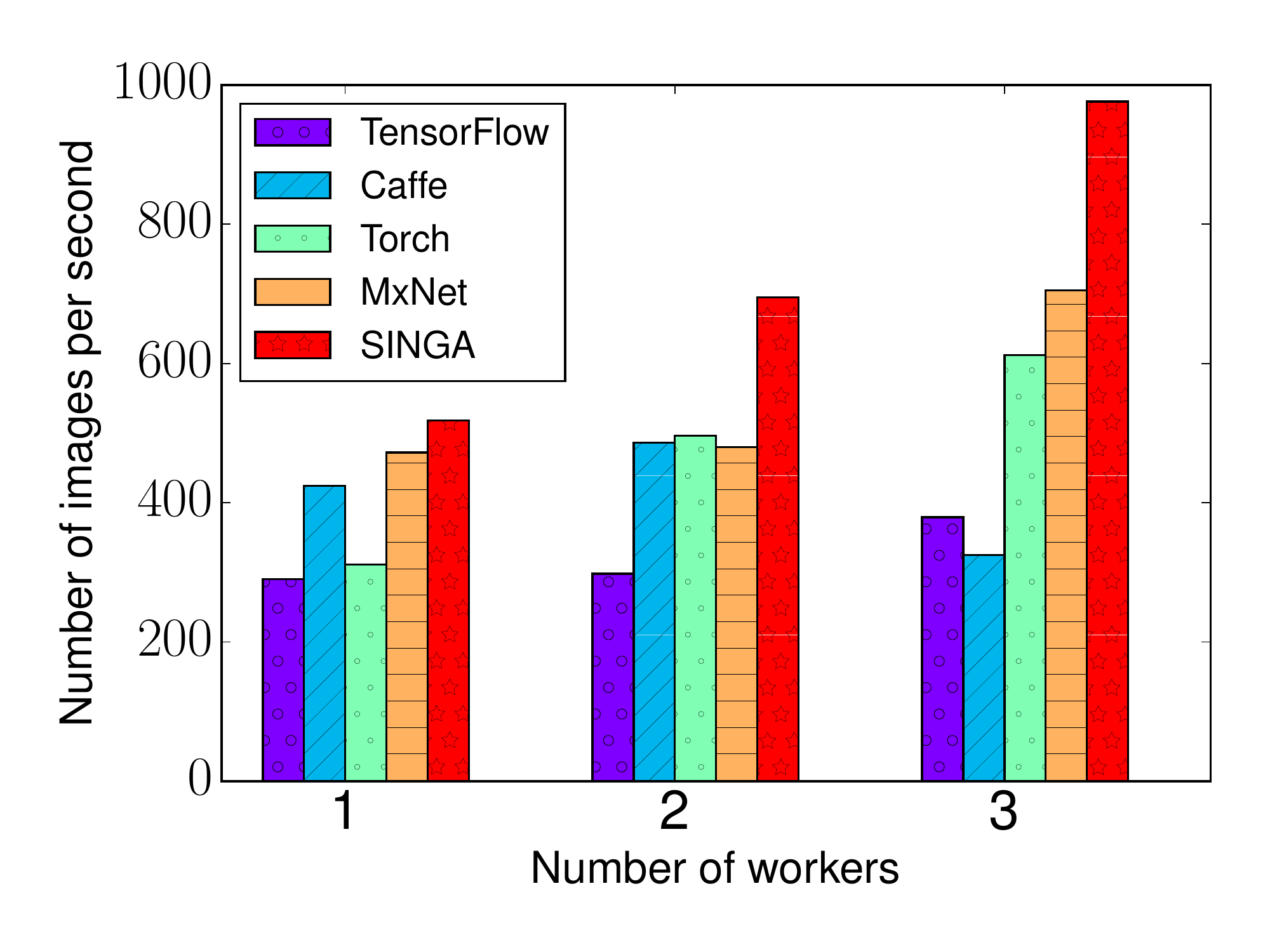}}
	\subfigure[\label{fig:fix-total} Total mini-batch size =288.]{ %
		\includegraphics[width=.37\textwidth]{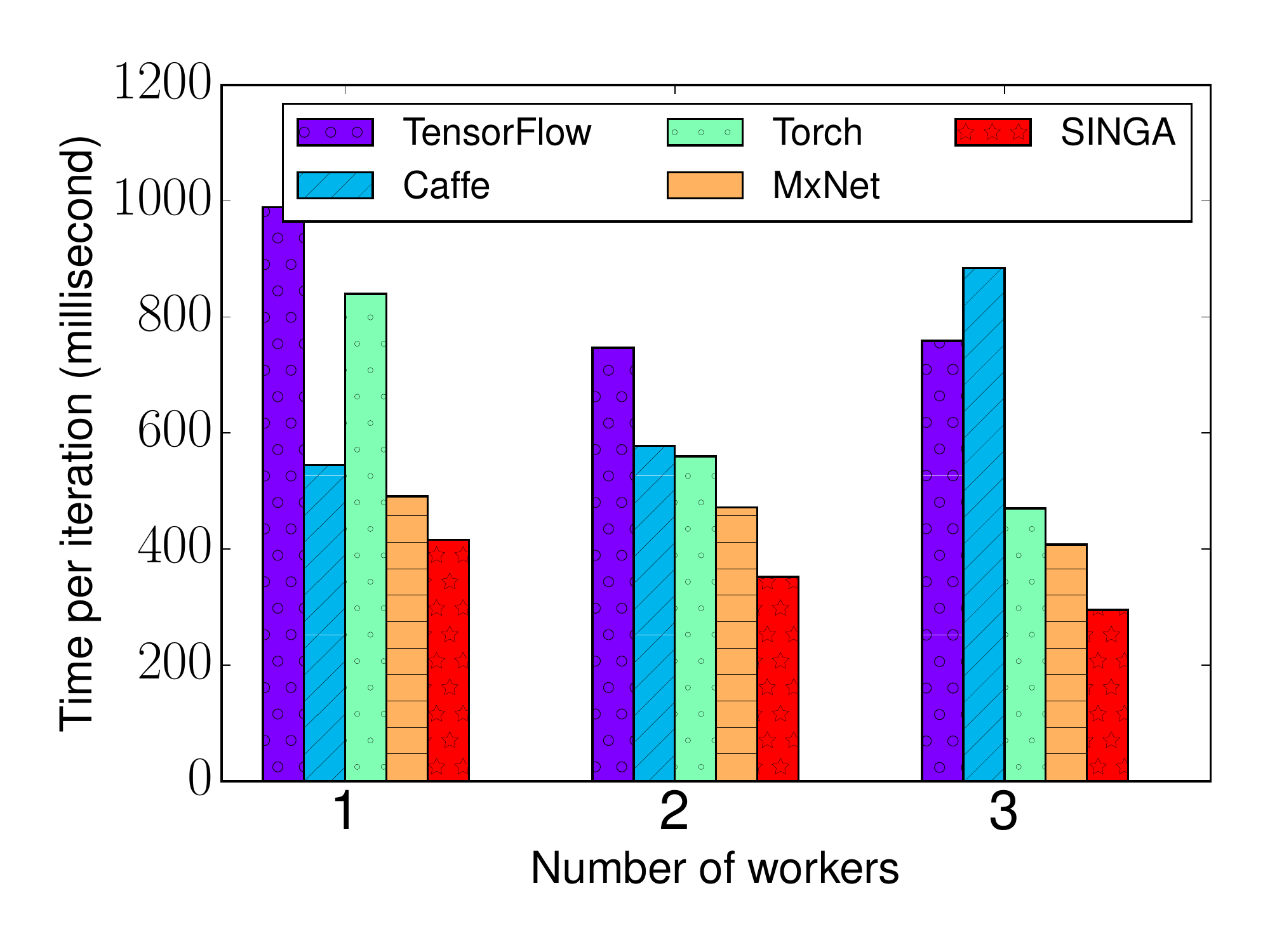}}
	\caption{Performance comparison of open source systems.}
\end{figure}

We first fixed the mini-batch size for each worker to be 96, and compared the five systems in terms
of throughput. Next, we fixed the overall mini-batch size to be $288$, i.e. the training workload per iteration is
fixed with each worker having a mini-batch size of $\frac{288}{n}$ (where $n$ is the number of workers), and compared
the five systems in terms of efficiency. For both sets of experiments, we varied the number of workers from $1$ to $3$.
The results are shown in Figure~\ref{fig:fix-per-worker} and Figure~\ref{fig:fix-total}. We can see that SINGA
outperforms other systems in both sets of experiments. On a single GPU, the difference with Torch and MxNet is not
significant because they also use cuDNN for the underlying computation of convolution, pooling etc. layers. The performance of TensorFlow, Caffe and Torch is consistent with that from the on-line benchmark site. On multiple
GPUs, SINGA has better performance thanks to the optimization techniques introduced in Section~\ref{sec:opt}. The performance of Caffe decreases when the number of workers is increased from 2 to 3. This could be caused by the tree reduction strategy\footnote{\url{https://github.com/BVLC/caffe/blob/master/docs/multigpu.md}} which works well for GPU cards with direct transferring capability. However, for our experiment, the GTX 970 cards do not provide such functionality, hence the data has to go through the CPU memory which incurs extra overhead when there are more than 2 workers. 

\section{Conclusion}\label{sec:conclusion}
In this paper, we proposed a distributed deep learning platform, called SINGA, for supporting multimedia applications.  SINGA offers a simple and intuitive programming model, making it accessible to even non-experts. SINGA is extensible and able to support a wide range of multimedia applications requiring  different deep learning models.  The flexible training architecture gives the user the chance to balance the trade-off between the training efficiency and convergence rate. Optimization techniques are applied to improve the training performance. We demonstrated the use of SINGA for representative multimedia applications using a CPU cluster and a single node with multiple GPU cards, and showed that the platform is both usable and scalable.

\begin{acks}
This work was in part supported by the National Research
Foundation, Prime Minister's Office, Singapore under its
Competitive Research Programme (CRP Award No. NRF-CRP8-2011-08) and A*STAR
project 1321202073. Gang Chen's work was supported by National Natural Science Foundation of
China (NSFC) Grant No. 61472348. We would like to thank the SINGA team members and NetEase
for their contributions to the implementation of the Apache SINGA system, and
the anonymous reviewers for their insightful and constructive comments.

\end{acks}

\bibliographystyle{ACM-Reference-Format-Journals}
\bibliography{paper}


\begin{thebibliography}{00}


\ifx \showCODEN    \undefined \def \showCODEN     #1{\unskip}     \fi
\ifx \showDOI      \undefined \def \showDOI       #1{{\tt DOI:}\penalty0{#1}\ }
  \fi
\ifx \showISBNx    \undefined \def \showISBNx     #1{\unskip}     \fi
\ifx \showISBNxiii \undefined \def \showISBNxiii  #1{\unskip}     \fi
\ifx \showISSN     \undefined \def \showISSN      #1{\unskip}     \fi
\ifx \showLCCN     \undefined \def \showLCCN      #1{\unskip}     \fi
\ifx \shownote     \undefined \def \shownote      #1{#1}          \fi
\ifx \showarticletitle \undefined \def \showarticletitle #1{#1}   \fi
\ifx \showURL      \undefined \def \showURL       #1{#1}          \fi

\bibitem[\protect\citeauthoryear{Abadi, Agarwal, Barham, Brevdo, Chen, Citro,
  Corrado, Davis, Dean, Devin, Ghemawat, Goodfellow, Harp, Irving, Isard, Jia,
  Jozefowicz, Kaiser, Kudlur, Levenberg, Man\'{e}, Monga, Moore, Murray, Olah,
  Schuster, Shlens, Steiner, Sutskever, Talwar, Tucker, Vanhoucke, Vasudevan,
  Vi\'{e}gas, Vinyals, Warden, Wattenberg, Wicke, Yu, and Zheng}{Abadi
  et~al\mbox{.}}{2015}]%
        {tensorflow2015-whitepaper}
{Mart\'{\i}n Abadi}, {Ashish Agarwal}, {Paul Barham}, {Eugene Brevdo}, {Zhifeng
  Chen}, {Craig Citro}, {Greg~S. Corrado}, {Andy Davis}, {Jeffrey Dean},
  {Matthieu Devin}, {Sanjay Ghemawat}, {Ian Goodfellow}, {Andrew Harp},
  {Geoffrey Irving}, {Michael Isard}, {Yangqing Jia}, {Rafal Jozefowicz},
  {Lukasz Kaiser}, {Manjunath Kudlur}, {Josh Levenberg}, {Dan Man\'{e}}, {Rajat
  Monga}, {Sherry Moore}, {Derek Murray}, {Chris Olah}, {Mike Schuster},
  {Jonathon Shlens}, {Benoit Steiner}, {Ilya Sutskever}, {Kunal Talwar}, {Paul
  Tucker}, {Vincent Vanhoucke}, {Vijay Vasudevan}, {Fernanda Vi\'{e}gas},
  {Oriol Vinyals}, {Pete Warden}, {Martin Wattenberg}, {Martin Wicke}, {Yuan
  Yu}, {and} {Xiaoqiang Zheng}. 2015.
\newblock {TensorFlow}: Large-Scale Machine Learning on Heterogeneous Systems.
\newblock   (2015).
\newblock
\showURL{%
\url{http://tensorflow.org/}}
\newblock
\shownote{Software available from tensorflow.org.}


\bibitem[\protect\citeauthoryear{Bastien, Lamblin, Pascanu, Bergstra,
  Goodfellow, Bergeron, Bouchard, and Bengio}{Bastien et~al\mbox{.}}{2012}]%
        {Bastien-Theano-2012}
{Fr{\'{e}}d{\'{e}}ric Bastien}, {Pascal Lamblin}, {Razvan Pascanu}, {James
  Bergstra}, {Ian~J. Goodfellow}, {Arnaud Bergeron}, {Nicolas Bouchard}, {and}
  {Yoshua Bengio}. 2012.
\newblock Theano: new features and speed improvements.
\newblock Deep Learning and Unsupervised Feature Learning NIPS 2012 Workshop.
  (2012).
\newblock


\bibitem[\protect\citeauthoryear{Chen, Li, Li, Lin, Wang, Wang, Xiao, Xu,
  Zhang, and Zhang}{Chen et~al\mbox{.}}{2015}]%
        {chen2015mxnet}
{Tianqi Chen}, {Mu Li}, {Yutian Li}, {Min Lin}, {Naiyan Wang}, {Minjie Wang},
  {Tianjun Xiao}, {Bing Xu}, {Chiyuan Zhang}, {and} {Zheng Zhang}. 2015.
\newblock \showarticletitle{MXNet: A Flexible and Efficient Machine Learning
  Library for Heterogeneous Distributed Systems}.
\newblock {\em arXiv preprint arXiv:1512.01274\/} (2015).
\newblock


\bibitem[\protect\citeauthoryear{Chilimbi, Suzue, Apacible, and
  Kalyanaraman}{Chilimbi et~al\mbox{.}}{2014}]%
        {186212}
{Trishul Chilimbi}, {Yutaka Suzue}, {Johnson Apacible}, {and} {Karthik
  Kalyanaraman}. 2014.
\newblock \showarticletitle{Project Adam: Building an Efficient and Scalable
  Deep Learning Training System}. In {\em OSDI}. USENIX Association, 571--582.
\newblock
\showISBNx{978-1-931971-16-4}
\showURL{%
\url{https://www.usenix.org/conference/osdi14/technical-sessions/presentation/chilimbi}}


\bibitem[\protect\citeauthoryear{Chua, Tang, Hong, Li, Luo, and Zheng}{Chua
  et~al\mbox{.}}{2009}]%
        {nus-wide-civr09}
{Tat-Seng Chua}, {Jinhui Tang}, {Richang Hong}, {Haojie Li}, {Zhiping Luo},
  {and} {Yan-Tao. Zheng}. July 8-10, 2009.
\newblock \showarticletitle{{NUS-WIDE}: A Real-World Web Image Database from
  {National University of Singapore}}. In {\em CIVR'09}.
\newblock


\bibitem[\protect\citeauthoryear{Ciresan, Meier, Gambardella, and
  Schmidhuber}{Ciresan et~al\mbox{.}}{2010}]%
        {DBLP:journals/corr/abs-1003-0358}
{Dan~Claudiu Ciresan}, {Ueli Meier}, {Luca~Maria Gambardella}, {and}
  {J{\"u}rgen Schmidhuber}. 2010.
\newblock \showarticletitle{Deep Big Simple Neural Nets Excel on Handwritten
  Digit Recognition}.
\newblock {\em CoRR\/}  {abs/1003.0358} (2010).
\newblock


\bibitem[\protect\citeauthoryear{Coates, Huval, Wang, Wu, Catanzaro, and
  Ng}{Coates et~al\mbox{.}}{2013}]%
        {DBLP:conf/icml/CoatesHWWCN13}
{Adam Coates}, {Brody Huval}, {Tao Wang}, {David~J. Wu}, {Bryan~C. Catanzaro},
  {and} {Andrew~Y. Ng}. 2013.
\newblock \showarticletitle{Deep learning with COTS HPC systems}. In {\em ICML
  (3)}. 1337--1345.
\newblock


\bibitem[\protect\citeauthoryear{Collobert, Kavukcuoglu, and Farabet}{Collobert
  et~al\mbox{.}}{2011}]%
        {collobert:2011c}
{R. Collobert}, {K. Kavukcuoglu}, {and} {C. Farabet}. 2011.
\newblock \showarticletitle{Torch7: A Matlab-like Environment for Machine
  Learning}. In {\em BigLearn, NIPS Workshop}.
\newblock


\bibitem[\protect\citeauthoryear{Dai, Wei, Zheng, Kim, Lee, Yin, Ho, and
  Xing}{Dai et~al\mbox{.}}{2013}]%
        {DBLP:journals/corr/DaiWZKLYHX13}
{Wei Dai}, {Jinliang Wei}, {Xun Zheng}, {Jin~Kyu Kim}, {Seunghak Lee}, {Junming
  Yin}, {Qirong Ho}, {and} {Eric~P. Xing}. 2013.
\newblock \showarticletitle{Petuum: {A} Framework for Iterative-Convergent
  Distributed {ML}}.
\newblock {\em CoRR\/}  {abs/1312.7651} (2013).
\newblock
\showURL{%
\url{http://arxiv.org/abs/1312.7651}}


\bibitem[\protect\citeauthoryear{Dean, Corrado, Monga, Chen, Devin, Le, Mao,
  Ranzato, Senior, Tucker, Yang, and Ng}{Dean et~al\mbox{.}}{2012}]%
        {DBLP:conf/nips/DeanCMCDLMRSTYN12}
{Jeffrey Dean}, {Greg Corrado}, {Rajat Monga}, {Kai Chen}, {Matthieu Devin},
  {Quoc~V. Le}, {Mark~Z. Mao}, {Marc'Aurelio Ranzato}, {Andrew~W. Senior},
  {Paul~A. Tucker}, {Ke Yang}, {and} {Andrew~Y. Ng}. 2012.
\newblock \showarticletitle{Large Scale Distributed Deep Networks}. In {\em
  NIPS}. 1232--1240.
\newblock


\bibitem[\protect\citeauthoryear{Dean and Ghemawat}{Dean and Ghemawat}{2004}]%
        {DBLP:conf/osdi/DeanG04}
{Jeffrey Dean} {and} {Sanjay Ghemawat}. 2004.
\newblock \showarticletitle{MapReduce: Simplified Data Processing on Large
  Clusters}. In {\em {(OSDI} 2004), San Francisco, California, USA, December
  6-8, 2004}. 137--150.
\newblock
\showURL{%
\url{http://www.usenix.org/events/osdi04/tech/dean.html}}


\bibitem[\protect\citeauthoryear{Duchi, Hazan, and Singer}{Duchi
  et~al\mbox{.}}{2011}]%
        {DBLP:journals/jmlr/DuchiHS11}
{John~C. Duchi}, {Elad Hazan}, {and} {Yoram Singer}. 2011.
\newblock \showarticletitle{Adaptive Subgradient Methods for Online Learning
  and Stochastic Optimization}.
\newblock {\em Journal of Machine Learning Research\/}  {12} (2011),
  2121--2159.
\newblock
\showURL{%
\url{http://dl.acm.org/citation.cfm?id=2021068}}


\bibitem[\protect\citeauthoryear{Feng, Wang, and Li}{Feng
  et~al\mbox{.}}{2014}]%
        {DBLP:conf/mm/FengWL14}
{Fangxiang Feng}, {Xiaojie Wang}, {and} {Ruifan Li}. 2014.
\newblock \showarticletitle{Cross-modal Retrieval with Correspondence
  Autoencoder}. In {\em ACM Multimedia}. 7--16.
\newblock
\showDOI{%
\url{http://dx.doi.org/10.1145/2647868.2654902}}


\bibitem[\protect\citeauthoryear{He, Zhang, Ren, and Sun}{He
  et~al\mbox{.}}{2015}]%
        {he2015deep}
{Kaiming He}, {Xiangyu Zhang}, {Shaoqing Ren}, {and} {Jian Sun}. 2015.
\newblock \showarticletitle{Deep Residual Learning for Image Recognition}.
\newblock {\em arXiv preprint arXiv:1512.03385\/} (2015).
\newblock


\bibitem[\protect\citeauthoryear{Hinton and Salakhutdinov}{Hinton and
  Salakhutdinov}{2006}]%
        {HinSal06}
{Geoffrey Hinton} {and} {Ruslan Salakhutdinov}. 2006.
\newblock \showarticletitle{Reducing the Dimensionality of Data with Neural
  Networks}.
\newblock {\em Science\/} {313}, 5786 (2006), 504 -- 507.
\newblock


\bibitem[\protect\citeauthoryear{Jia, Shelhamer, Donahue, Karayev, Long,
  Girshick, Guadarrama, and Darrell}{Jia et~al\mbox{.}}{2014}]%
        {jia2014caffe}
{Yangqing Jia}, {Evan Shelhamer}, {Jeff Donahue}, {Sergey Karayev}, {Jonathan
  Long}, {Ross Girshick}, {Sergio Guadarrama}, {and} {Trevor Darrell}. 2014.
\newblock \showarticletitle{Caffe: Convolutional Architecture for Fast Feature
  Embedding}.
\newblock {\em arXiv preprint arXiv:1408.5093\/} (2014).
\newblock


\bibitem[\protect\citeauthoryear{Jiang, Chen, Ooi, Tan, and Wu}{Jiang
  et~al\mbox{.}}{2014}]%
        {DBLP:journals/pvldb/Jiang0OTW14}
{Dawei Jiang}, {Gang Chen}, {Beng~Chin Ooi}, {Kian{-}Lee Tan}, {and} {Sai Wu}.
  2014.
\newblock \showarticletitle{{epiC}: an Extensible and Scalable System for
  Processing Big Data}.
\newblock {\em {PVLDB}\/} {7}, 7 (2014), 541--552.
\newblock
\showURL{%
\url{http://www.vldb.org/pvldb/vol7/p541-jiang.pdf}}


\bibitem[\protect\citeauthoryear{Krizhevsky}{Krizhevsky}{2014}]%
        {DBLP:journals/corr/Krizhevsky14}
{Alex Krizhevsky}. 2014.
\newblock \showarticletitle{One weird trick for parallelizing convolutional
  neural networks}.
\newblock {\em CoRR\/}  {abs/1404.5997} (2014).
\newblock


\bibitem[\protect\citeauthoryear{Krizhevsky, Sutskever, and Hinton}{Krizhevsky
  et~al\mbox{.}}{2012}]%
        {DBLP:conf/nips/KrizhevskySH12}
{Alex Krizhevsky}, {Ilya Sutskever}, {and} {Geoffrey~E. Hinton}. 2012.
\newblock \showarticletitle{ImageNet Classification with Deep Convolutional
  Neural Networks}. In {\em NIPS}. 1106--1114.
\newblock


\bibitem[\protect\citeauthoryear{Le, Ranzato, Monga, Devin, Corrado, Chen,
  Dean, and Ng}{Le et~al\mbox{.}}{2012}]%
        {DBLP:conf/icml/LeRMDCCDN12}
{Quoc~V. Le}, {Marc'Aurelio Ranzato}, {Rajat Monga}, {Matthieu Devin}, {Greg
  Corrado}, {Kai Chen}, {Jeffrey Dean}, {and} {Andrew~Y. Ng}. 2012.
\newblock \showarticletitle{Building high-level features using large scale
  unsupervised learning}. In {\em ICML}.
\newblock


\bibitem[\protect\citeauthoryear{LeCun, Bottou, Orr, and M{\"{u}}ller}{LeCun
  et~al\mbox{.}}{1996}]%
        {DBLP:conf/nips/LeCunBOM96}
{Yann LeCun}, {L{\'{e}}on Bottou}, {Genevieve~B. Orr}, {and} {Klaus{-}Robert
  M{\"{u}}ller}. 1996.
\newblock \showarticletitle{Effiicient BackProp}. In {\em Neural Networks:
  Tricks of the Trade}. 9--50.
\newblock
\showDOI{%
\url{http://dx.doi.org/10.1007/3-540-49430-8_2}}


\bibitem[\protect\citeauthoryear{Mikolov, Kombrink, Burget, Cernock{\'{y}}, and
  Khudanpur}{Mikolov et~al\mbox{.}}{2011}]%
        {DBLP:conf/icassp/MikolovKBCK11}
{Tomas Mikolov}, {Stefan Kombrink}, {Luk{\'{a}}s Burget}, {Jan Cernock{\'{y}}},
  {and} {Sanjeev Khudanpur}. 2011.
\newblock \showarticletitle{Extensions of recurrent neural network language
  model}. In {\em ICASSP}. {IEEE}, 5528--5531.
\newblock
\showDOI{%
\url{http://dx.doi.org/10.1109/ICASSP.2011.5947611}}


\bibitem[\protect\citeauthoryear{Mikolov, Sutskever, Chen, Corrado, and
  Dean}{Mikolov et~al\mbox{.}}{2013}]%
        {DBLP:conf/nips/MikolovSCCD13}
{Tomas Mikolov}, {Ilya Sutskever}, {Kai Chen}, {Gregory~S. Corrado}, {and}
  {Jeffrey Dean}. 2013.
\newblock \showarticletitle{Distributed Representations of Words and Phrases
  and their Compositionality}. In {\em NIPS}. 3111--3119.
\newblock


\bibitem[\protect\citeauthoryear{Ooi, Tan, Wang, Wang, Cai, Chen, Gao, Luo,
  Tung, Wang, Xie, Zhang, and Zheng}{Ooi et~al\mbox{.}}{2015}]%
        {singa-oss}
{Beng~Chin Ooi}, {Kian-Lee Tan}, {Sheng Wang}, {Wei Wang}, {Qingchao Cai},
  {Gang Chen}, {Jinyang Gao}, {Zhaojing Luo}, {Anthony K.~H. Tung}, {Yuan
  Wang}, {Zhongle Xie}, {Meihui Zhang}, {and} {Kaiping Zheng}. 2015.
\newblock \showarticletitle{{SINGA}: A Distributed Deep Learning Platform}. In
  {\em ACM Multimedia}.
\newblock


\bibitem[\protect\citeauthoryear{Paine, Jin, Yang, Lin, and Huang}{Paine
  et~al\mbox{.}}{2013}]%
        {DBLP:journals/corr/PaineJYLH13}
{Thomas Paine}, {Hailin Jin}, {Jianchao Yang}, {Zhe Lin}, {and} {Thomas~S.
  Huang}. 2013.
\newblock \showarticletitle{{GPU} Asynchronous Stochastic Gradient Descent to
  Speed Up Neural Network Training}.
\newblock {\em CoRR\/}  {abs/1312.6186} (2013).
\newblock


\bibitem[\protect\citeauthoryear{Recht, Re, Wright, and Niu}{Recht
  et~al\mbox{.}}{2011}]%
        {DBLP:conf/nips/RechtRWN11}
{Benjamin Recht}, {Christopher Re}, {Stephen~J. Wright}, {and} {Feng Niu}.
  2011.
\newblock \showarticletitle{Hogwild: A Lock-Free Approach to Parallelizing
  Stochastic Gradient Descent}. In {\em NIPS}. 693--701.
\newblock


\bibitem[\protect\citeauthoryear{Seide, Fu, Droppo, Li, and Yu}{Seide
  et~al\mbox{.}}{2014}]%
        {DBLP:conf/interspeech/SeideFDLY14}
{Frank Seide}, {Hao Fu}, {Jasha Droppo}, {Gang Li}, {and} {Dong Yu}. 2014.
\newblock \showarticletitle{1-bit stochastic gradient descent and its
  application to data-parallel distributed training of speech DNNs}. In {\em
  {INTERSPEECH} 2014, 15th Annual Conference of the International Speech
  Communication Association, Singapore, September 14-18, 2014}. 1058--1062.
\newblock


\bibitem[\protect\citeauthoryear{Shen, Ooi, and Tan}{Shen
  et~al\mbox{.}}{2000}]%
        {DBLP:conf/mm/ShenOT00}
{Heng~Tao Shen}, {Beng~Chin Ooi}, {and} {Kian{-}Lee Tan}. 2000.
\newblock \showarticletitle{Giving meanings to {WWW} images}. In {\em ACM
  Multimedia}. 39--47.
\newblock


\bibitem[\protect\citeauthoryear{Simonyan and Zisserman}{Simonyan and
  Zisserman}{2014}]%
        {DBLP:journals/corr/SimonyanZ14a}
{Karen Simonyan} {and} {Andrew Zisserman}. 2014.
\newblock \showarticletitle{Very Deep Convolutional Networks for Large-Scale
  Image Recognition}.
\newblock {\em CoRR\/}  {abs/1409.1556} (2014).
\newblock
\showURL{%
\url{http://arxiv.org/abs/1409.1556}}


\bibitem[\protect\citeauthoryear{Szegedy, Liu, Jia, Sermanet, Reed, Anguelov,
  Erhan, Vanhoucke, and Rabinovich}{Szegedy et~al\mbox{.}}{2014}]%
        {DBLP:journals/corr/SzegedyLJSRAEVR14}
{Christian Szegedy}, {Wei Liu}, {Yangqing Jia}, {Pierre Sermanet}, {Scott
  Reed}, {Dragomir Anguelov}, {Dumitru Erhan}, {Vincent Vanhoucke}, {and}
  {Andrew Rabinovich}. 2014.
\newblock \showarticletitle{Going Deeper with Convolutions}.
\newblock {\em CoRR\/}  {abs/1409.4842} (2014).
\newblock


\bibitem[\protect\citeauthoryear{Wan, Wang, Hoi, Wu, Zhu, Zhang, and Li}{Wan
  et~al\mbox{.}}{2014}]%
        {DBLP:conf/mm/WanWHWZZL14}
{Ji Wan}, {Dayong Wang}, {Steven Chu~Hong Hoi}, {Pengcheng Wu}, {Jianke Zhu},
  {Yongdong Zhang}, {and} {Jintao Li}. 2014.
\newblock \showarticletitle{Deep Learning for Content-Based Image Retrieval:
  {A} Comprehensive Study}. In {\em ACM Multimedia}. 157--166.
\newblock


\bibitem[\protect\citeauthoryear{Wang, Chen, Dinh, Gao, Ooi, Tan, and
  Wang}{Wang et~al\mbox{.}}{2015}]%
        {singa-mm}
{Wei Wang}, {Gang Chen}, {Tien Tuan~Anh Dinh}, {Jinyang Gao}, {Beng~Chin Ooi},
  {Kian-Lee Tan}, {and} {Sheng Wang}. 2015.
\newblock \showarticletitle{{SINGA}: Putting Deep Learning in the Hands of
  Multimedia Users}. In {\em ACM Multimedia}.
\newblock


\bibitem[\protect\citeauthoryear{Wang, Ooi, Yang, Zhang, and Zhuang}{Wang
  et~al\mbox{.}}{2014}]%
        {DBLP:journals/pvldb/WangOYZZ14}
{Wei Wang}, {Beng~Chin Ooi}, {Xiaoyan Yang}, {Dongxiang Zhang}, {and} {Yueting
  Zhuang}. 2014.
\newblock \showarticletitle{Effective Multi-Modal Retrieval based on Stacked
  Auto-Encoders}.
\newblock {\em {PVLDB}\/} {7}, 8 (2014), 649--660.
\newblock


\bibitem[\protect\citeauthoryear{Wang, Yang, Ooi, Zhang, and Zhuang}{Wang
  et~al\mbox{.}}{2015}]%
        {raey}
{Wei Wang}, {Xiaoyan Yang}, {Beng~Chin Ooi}, {Dongxiang Zhang}, {and} {Yueting
  Zhuang}. 2015.
\newblock \showarticletitle{Effective deep learning-based multi-modal
  retrieval}.
\newblock {\em The VLDB Journal\/} (2015), 1--23.
\newblock
\showISSN{1066-8888}
\showDOI{%
\url{http://dx.doi.org/10.1007/s00778-015-0391-4}}


\bibitem[\protect\citeauthoryear{Wang and Wang}{Wang and Wang}{2014}]%
        {DBLP:conf/mm/WangW14}
{Xinxi Wang} {and} {Ye Wang}. 2014.
\newblock \showarticletitle{Improving Content-based and Hybrid Music
  Recommendation using Deep Learning}. In {\em ACM Multimedia}. 627--636.
\newblock
\showDOI{%
\url{http://dx.doi.org/10.1145/2647868.2654940}}


\bibitem[\protect\citeauthoryear{Wu, Yan, Shan, Dang, and Sun}{Wu
  et~al\mbox{.}}{2015}]%
        {DBLP:journals/corr/WuYSDS15}
{Ren Wu}, {Shengen Yan}, {Yi Shan}, {Qingqing Dang}, {and} {Gang Sun}. 2015.
\newblock \showarticletitle{Deep Image: Scaling up Image Recognition}.
\newblock {\em CoRR\/}  {abs/1501.02876} (2015).
\newblock
\showURL{%
\url{http://arxiv.org/abs/1501.02876}}


\bibitem[\protect\citeauthoryear{Wu, Jiang, Wang, Pu, and Xue}{Wu
  et~al\mbox{.}}{2014}]%
        {DBLP:conf/mm/WuJWPX14}
{Zuxuan Wu}, {Yu{-}Gang Jiang}, {Jun Wang}, {Jian Pu}, {and} {Xiangyang Xue}.
  2014.
\newblock \showarticletitle{Exploring Inter-feature and Inter-class
  Relationships with Deep Neural Networks for Video Classification}. In {\em
  ACM Multimedia}. 167--176.
\newblock
\showDOI{%
\url{http://dx.doi.org/10.1145/2647868.2654931}}


\bibitem[\protect\citeauthoryear{Yadan, Adams, Taigman, and Ranzato}{Yadan
  et~al\mbox{.}}{2013}]%
        {DBLP:journals/corr/YadanATR13}
{Omry Yadan}, {Keith Adams}, {Yaniv Taigman}, {and} {Marc'Aurelio Ranzato}.
  2013.
\newblock \showarticletitle{Multi-{GPU} Training of ConvNets}.
\newblock {\em CoRR\/}  {abs/1312.5853} (2013).
\newblock


\bibitem[\protect\citeauthoryear{You, Luo, Jin, and Yang}{You
  et~al\mbox{.}}{2015}]%
        {DBLP:conf/mm/YouLJY15}
{Quanzeng You}, {Jiebo Luo}, {Hailin Jin}, {and} {Jianchao Yang}. 2015.
\newblock \showarticletitle{Joint Visual-Textual Sentiment Analysis with Deep
  Neural Networks}. In {\em Proceedings of the 23rd Annual {ACM} Conference on
  Multimedia Conference, {MM} '15, Brisbane, Australia, October 26 - 30, 2015}.
  1071--1074.
\newblock
\showDOI{%
\url{http://dx.doi.org/10.1145/2733373.2806284}}


\bibitem[\protect\citeauthoryear{Zaharia, Chowdhury, Das, Dave, Ma, McCauly,
  Franklin, Shenker, and Stoica}{Zaharia et~al\mbox{.}}{2012}]%
        {DBLP:conf/nsdi/ZahariaCDDMMFSS12}
{Matei Zaharia}, {Mosharaf Chowdhury}, {Tathagata Das}, {Ankur Dave}, {Justin
  Ma}, {Murphy McCauly}, {Michael~J. Franklin}, {Scott Shenker}, {and} {Ion
  Stoica}. 2012.
\newblock \showarticletitle{Resilient Distributed Datasets: A Fault-Tolerant
  Abstraction for In-Memory Cluster Computing}. In {\em NSDI}. 15--28.
\newblock


\bibitem[\protect\citeauthoryear{Zhang and Re}{Zhang and Re}{2014}]%
        {DBLP:dblp_journals/corr/ZhangR14}
{Ce Zhang} {and} {Christopher Re}. 2014.
\newblock \showarticletitle{DimmWitted: {A} Study of Main-Memory Statistical
  Analytics}.
\newblock {\em {PVLDB}\/} {7}, 12 (2014), 1283--1294.
\newblock
\showURL{%
\url{http://www.vldb.org/pvldb/vol7/p1283-zhang.pdf}}


\bibitem[\protect\citeauthoryear{Zhang, Yang, Luan, Yang, and Chua}{Zhang
  et~al\mbox{.}}{2014}]%
        {DBLP:conf/mm/ZhangYLYC14}
{Hanwang Zhang}, {Yang Yang}, {Huan{-}Bo Luan}, {Shuicheng Yang}, {and}
  {Tat{-}Seng Chua}. 2014.
\newblock \showarticletitle{Start from Scratch: Towards Automatically
  Identifying, Modeling, and Naming Visual Attributes}. In {\em ACM
  Multimedia}. 187--196.
\newblock


\end{thebibliography}




\end{document}